\documentclass{article}

\usepackage[main, final]{neurips_2026}

\usepackage[utf8]{inputenc}      
\usepackage[T1]{fontenc}
\usepackage{hyperref}       
\usepackage{url}            
\usepackage{booktabs}       
\usepackage{amsmath}        
\usepackage{amsfonts}       
\usepackage{nicefrac}       
\usepackage{microtype}      
\usepackage[table]{xcolor}  
\usepackage{textcomp}       
\usepackage{amssymb}        
\usepackage{multirow}
\usepackage{makecell}
\usepackage{float} 
\usepackage{graphicx}
\usepackage{wrapfig}
\usepackage{algorithm}
\usepackage{algorithmic}
\usepackage{tabularx}
\usepackage[table]{xcolor}
\usepackage{booktabs}
\usepackage{wrapfig}
\usepackage[most]{tcolorbox}
\usepackage{xcolor}
\usepackage{caption} 

\tcbset{
  gencommon/.style={
    enhanced,
    colback=white,
    colframe=black!25,
    colbacktitle=black!8,
    coltitle=black,
    boxrule=0.35pt,
    arc=0.8mm,
    left=5pt,
    right=5pt,
    top=5pt,
    bottom=5pt,
    before skip=0.7em,
    after skip=0.7em,
    fonttitle=\footnotesize\sffamily,
    fontupper=\footnotesize
  }
}

\newtcolorbox{genshort}[6][]{
  gencommon,
  before upper={
    \setlength{\parindent}{0pt}
    \setlength{\parskip}{0.35em}
    \textbf{Gen. PPL:} #5
    \quad
    \textbf{Entropy:} #6
    \par\smallskip\hrule\smallskip
  },
  #1
}

\newtcolorbox{genlong}[6][]{
  gencommon,
  breakable,
  title={\textbf{#2} --- unconditional sample, length #3},
  title after break={\textbf{#2} --- continued},
  before upper={
    \setlength{\parindent}{0pt}
    \setlength{\parskip}{0.35em}
    \raggedright
    \textbf{NFE:} #4
    \quad
    \textbf{Gen. PPL:} #5
    \quad
    \textbf{Entropy:} #6
    \par\smallskip\hrule\smallskip
  },
  #1
}
\usepackage[capitalize,noabbrev]{cleveref}  
\usepackage{tocloft}        

\definecolor{myyellow}{RGB}{222, 195, 23}
\definecolor{linkblue}{RGB}{25,95,125}
\hypersetup{
  colorlinks=true,
  linkcolor=linkblue,
  citecolor=linkblue,
  urlcolor=linkblue
}

\usepackage[most]{tcolorbox}

\newtcolorbox{samplebox}[1]{
	enhanced,
	colback=linkblue!5!white,
	colframe=black!60,
	boxrule=0.6pt,
	arc=6pt,
	left=4pt, right=4pt, top=4pt, bottom=4pt,
	boxsep=3pt,
	before upper={\noindent\textbf{\small #1}\par\smallskip},
}

\newtcolorbox{keyidea}{
  enhanced,
  colback=blue!2,
  colframe=blue!35!black,
  boxrule=0.45pt,
  arc=2pt,
  left=5pt,
  right=5pt,
  top=3pt,
  bottom=3pt,
  before skip=0.6em,
  after skip=0.6em
}

\title{How to Train Your Latent Diffusion Language Model Jointly With the Latent Space}

\newcommand{\affHSE}{HSE University, Moscow, Russia}
\newcommand{\affConstructor}{Constructor University, Bremen, Germany}
\newcommand{\affAAII}{Applied AI Institute, Moscow, Russia}
\newcommand{\affAXXX}{AXXX, Russia}
\newcommand{\affIndep}{Independent researcher}

\author{%
  Viacheslav Meshchaninov\textsuperscript{1},\;
  Alexander Shabalin\textsuperscript{1},\;
  Egor Chimbulatov\textsuperscript{2}, \\ \bf
  Nikita Gushchin\textsuperscript{3,4},\;
  Ilya Koziev\textsuperscript{5},\;
  Alexander Korotin\textsuperscript{3,4},\;
  Dmitry Vetrov\textsuperscript{1}
}

\begin{document}
\setcounter{tocdepth}{-1} 

\maketitle

{\let\thefootnotesave\thefootnote
 \renewcommand{\thefootnote}{}%
 \footnotetext{%
   \textsuperscript{1}\,\affConstructor;\quad
   \textsuperscript{2}\,\affHSE;\quad
   \textsuperscript{3}\,\affAAII;\quad
   \textsuperscript{4}\,\affAXXX;\quad
   \textsuperscript{5}\,\affIndep.\\
   \textsuperscript{*}\,Corresponding author:
   \texttt{vmeshchani@constructor.university},
   \texttt{meshchaninov.viacheslav@gmail.com}.}%
 \let\thefootnote\thefootnotesave}

\begin{abstract}
  Latent diffusion models offer an attractive alternative to discrete diffusion for non-autoregressive text generation by operating on continuous text representations and denoising entire sequences in parallel. The major challenge in latent diffusion modeling is constructing a suitable latent space. In this work, we present the \textbf{Latent Diffusion Language Model (LDLM)}, in which the latent encoder, diffusion model, and decoder are trained jointly. LDLM builds its latent space by reshaping the representations of a pre-trained language model with a trainable encoder, yielding latents that are easy to both denoise and decode into tokens. We show that naive joint training produces a low-quality diffusion model, and propose a simple training recipe consisting of an MSE decoder loss, diffusion-to-encoder warmup, adaptive timestep sampling, and decoder-input noise. Ablations show that each component substantially impacts generation performance. On OpenWebText and LM1B, LDLM achieves better generation performance than existing discrete and continuous diffusion language models while being $2{\text -}13\times$ faster, indicating that jointly learning the latent space is a key step toward making latent diffusion competitive for text generation.
  Our \href{https://github.com/MeshchaninovViacheslav/ldlm}{code} is publicly available.
\end{abstract}


\begin{figure}[H]
    \centering
    \begin{tabular}{c}
    \includegraphics[width=0.35\textwidth]{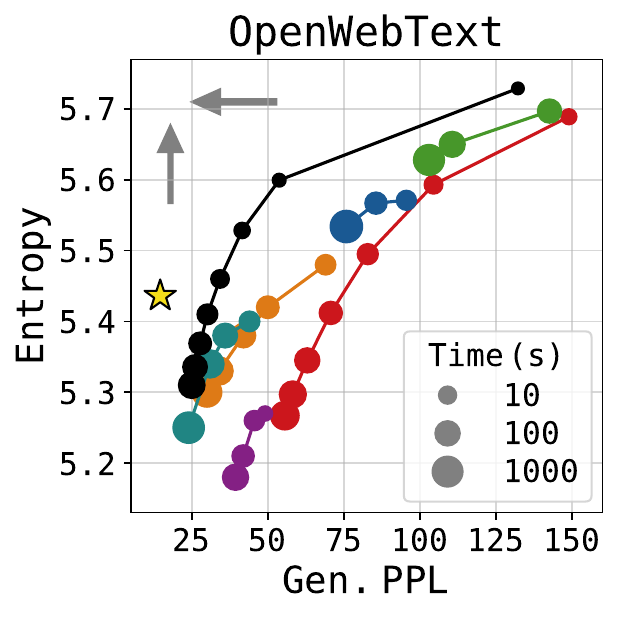}\hspace{15pt}
    \includegraphics[width=0.35\textwidth]{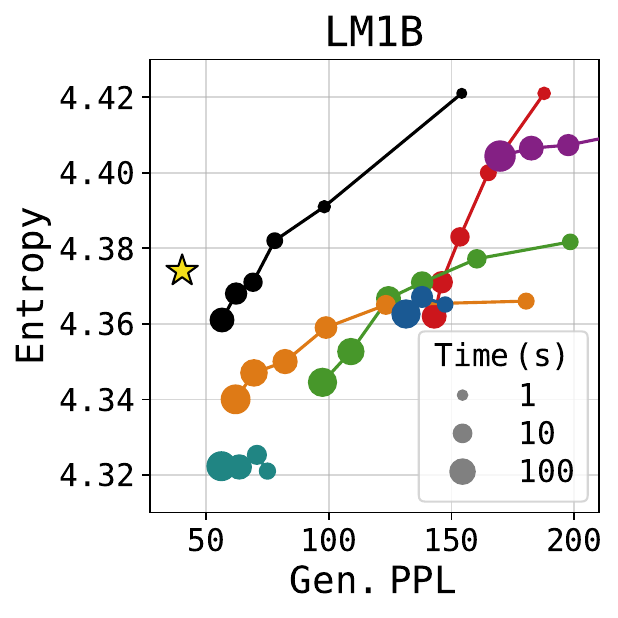}
    \\
    \includegraphics[width=0.55\textwidth]{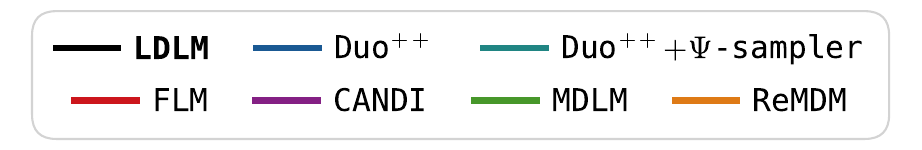}
    \end{tabular}
    \caption{\textbf{Quality-diversity trade-off in text generation.} Pareto frontiers are obtained by sweeping NFEs. The marker size denotes the generation time, and the \textcolor{myyellow}{yellow} star shows the statistics of real texts. The proposed \textbf{LDLM} achieves the best trade-off between Gen. PPL$\,(\downarrow)$ and entropy$\,(\uparrow)$ on both OpenWebText and LM1B, while remaining faster than competing baselines.
    }
    \label{fig:owt_lm1b}
\end{figure}

\section{Introduction}

Autoregressive models are the current standard for text generation~\citep{openai2024gpt4technicalreport, jiang2023mistral7b, deepseek}. However, despite their prevalence, they are constrained by their left-to-right generation pattern, which prevents them from correcting previous mistakes or generating more than one token at a time~\citep{ye2024beyond}.
Diffusion language models offer an alternative paradigm~\citep{diffusion-lm, cosmos, duo, candi}: they generate text through iterative refinement, updating all positions in parallel and providing greater control over the generated sequence.

Text diffusion models are commonly divided into discrete and continuous approaches.
Discrete diffusion operates directly in token space by corrupting and denoising categorical states, and has become the most developed direction for diffusion language modeling~\citep{mdlm, llada, mercury}.
However, discrete diffusion models suffer from the factorization of the joint token distribution, which causes all tokens to be predicted independently at each denoising step.
This limits their ability to generate multiple tokens simultaneously, making high-quality few-step generation difficult~\citep{wu2025fast, wang2025remasking, kang2026parallelbench}.

Continuous text diffusion models, in contrast, operate in a latent space and use a decoder to map the latents back to tokens only at the end of generation. This allows all positions to be refined gradually, avoiding the need to commit to a particular token before sampling is complete~\citep{diffusion-lm, sed}.
Continuous diffusion can be defined at different levels of representation.
One option is to apply Gaussian diffusion to simple token representations, such as shallow embeddings or one-hot encodings, and recent work shows that this approach can match strong discrete baselines~\citep{dinoiser, cdcd, lee2026flow}.
Going further, several works encode text using transformer-based encoders to build \emph{latent diffusion models}, which substantially improve the latent space and yield higher generation quality~\citep{ld4lg, planner, cosmos}.
This suggests that the structure of the latent space is the key component of text diffusion, significantly influencing generation quality.

To construct the best possible latent space, it is natural to \emph{train the encoder jointly with the diffusion model}.
However, while prior works have successfully done so for shallow embeddings~\citep{diffusion-lm, diffuseq, plaid}, joint training has never been implemented for transformer-based encoders: existing latent diffusion models rely on pre-trained encoders that remain frozen during diffusion training~\citep{tencdm, cosmos}.
As a result, shallow token embeddings yield a suboptimal latent space even when perfectly tuned for diffusion, while representations from pre-trained encoders are not optimally suited for the diffusion, since these encoders are trained on surrogate tasks.
In this paper, we address this gap. We show that jointly training a latent encoder and a diffusion model is non-trivial, and identify a set of simple techniques that make it work, achieving state-of-the-art generation quality.
Our contributions are:
\begin{itemize}
    \item We introduce \textbf{LDLM}, a latent diffusion language model that jointly trains the latent encoder, diffusion model, and decoder, directly shaping the latent space for diffusion (\cref{sec:method}).
    \item We show that naive joint training leads to low-diversity generations, and propose a simple recipe that makes end-to-end latent learning effective for text diffusion (\cref{sec:recipe,sec:ablations}).
    \item We compare text diffusion spaces and show that jointly learned contextual latents outperform alternative latent encoding approaches (\cref{sec:space_comparison}).
    \item We demonstrate on LM1B and OpenWebText that \textbf{LDLM} achieves higher performance than recent discrete and continuous diffusion language models, while staying $2{\text -}13\times$ faster (\cref{sec:main_results}).
\end{itemize}

\section{Related work}


\paragraph{Discrete diffusion models.}
Discrete diffusion has become the most developed approach for diffusion-based text generation. 
These methods define a noising process directly over categorical token variables. 
Most existing formulations use one of two terminal noised distributions: \emph{uniform} diffusion~\citep{duo}, where corrupted tokens approach the uniform distribution over the vocabulary, and \emph{masking} or absorbing diffusion~\citep{mdlm, md4}, where corrupted tokens are mapped to a dedicated mask token. 
While being more popular, masking-based models suffer from two additional limitations. 
First, once a token is unmasked, standard sampling cannot naturally revise it, necessitating explicit remasking strategies~\citep{wang2025remasking, meshchaninov2025guided} or edit operations~\citep{havasi2025edit, song2025seeddiffusionlargescalediffusion}. 
Second, although the model predicts distributions for all masked positions at each denoising step, standard samplers update only the subset selected for unmasking and discard the remaining predictions, even though these predictions can provide useful guidance~\citep{candi}.
Most importantly, standard discrete diffusion models parameterize each reverse step with a factorized per-position distribution. 
As a result, a denoising step consists of independent token-wise decisions rather than a joint sampling of tokens, making it difficult to update several tokens at once~\citep{codar, kang2026parallelbench}.

\paragraph{Continuous diffusion on token-level representations.}
An alternative line of work embeds tokens into a continuous space and applies Gaussian diffusion to these representations.
The token-level representations fall into two groups. The first uses fixed encodings such as one-hot vectors or points on the probability simplex~\citep{tess, potaptchik2026discrete, lee2026flow, roos2026categorical}.
The second employs token embeddings, usually learning the embedding matrix jointly with the diffusion model~\citep{diffusion-lm, diffuseq, han2022ssd}.
However, the joint embedding-diffusion optimization can be unstable and might require regularization to avoid collapse or norm explosion~\citep{sed, cdcd}. 
In general, token-level representations lack contextual semantic structure, making them a suboptimal choice for diffusion modeling~\citep{ld4lg, tencdm, shabalin2026gaussiandiffusionmodelsfail}.

\paragraph{Latent diffusion for text.} To address this, a third line of work constructs a sequence-level latent space from the contextual outputs of a pre-trained text encoder, which is frozen during diffusion training. Contextual information has been shown to substantially improve the quality of the latent space and, in turn, diffusion model performance~\citep{tencdm}.
Several methods train dedicated encoders to better shape the latent geometry: PLANNER~\citep{planner} and LD4LG~\citep{ld4lg} compress the encoder outputs through a lower-dimensional bottleneck, while COSMOS~\citep{cosmos} improves smoothness and robustness via masking and noise injection during autoencoder training.
Together, these works establish that the geometry of the latent space is critical to text diffusion performance.
However, in all of these methods the latent encoder is trained independently of the diffusion model.
Their training objectives are constructed to supplement the latent space with valuable properties such as smoothness and reconstruction fidelity, but they do not directly optimize for the needs of the diffusion model.
In contrast, our work \textbf{trains the latent encoder jointly with the diffusion model}, allowing the diffusion objective to directly shape its own latent space.


\section{Preliminaries}\label{sec:preliminaries}

\paragraph{Continuous diffusion.}
Continuous diffusion models~\citep{ddpm, song2020score} define a forward process that gradually corrupts a clean sample $\mathbf{z}_0 \sim p_{\mathrm{data}}$ with Gaussian noise, $\mathbf{z}_t = \sqrt{\bar{\alpha}_t} \mathbf{z}_0 + \sqrt{1 - \bar{\alpha}_t} \epsilon$, where $t \sim \mathcal{U}[0, 1], \epsilon \sim \mathcal{N}(\mathbf{0}, \mathbf{I})$ and $\bar{\alpha}_t$ defines the noise schedule which monotonically decreases with $t$ from $\bar{\alpha}_0 = 1$ to $\bar{\alpha}_1 = 0$.
A neural network $\hat{\mathbf{z}}_\theta(\mathbf{z}_t, t)$ is trained to recover $\mathbf{z}_0$ from the noisy observation $\mathbf{z}_t$ by minimizing
the denoising objective
\begin{equation}\label{eq:diff_loss}
  \mathcal{L}_{\mathrm{diff}}
  =
  \mathbb{E}_{\mathbf{z}_0,\, t,\, \epsilon}
  \bigl[
    \|\hat{\mathbf{z}}_\theta(\mathbf{z}_t, t) - \mathbf{z}_0\|^2
  \bigr].
\end{equation}

\paragraph{Continuous text diffusion.}
Let $\mathbf{w} = (w_1, \dots, w_n) \in \mathcal{V}^n$ be a discrete token sequence of
fixed length $n$, where $\mathcal{V}$ is the vocabulary. 
Since Gaussian diffusion is defined on continuous variables, the sequence is
first mapped into a continuous latent representation $\mathbf{z}_0 = \mathcal{E}(\mathbf{w})$, where $\mathcal{E}: \mathcal{V}^n \to \mathbb{R}^{n \times d}$ denotes a mapping function.
The diffusion model learns the denoising task in the obtained latent space by optimizing the loss in Eq.~(\ref{eq:diff_loss}).
To map the latent back into a sequence of tokens a decoding function is used, $\mathbf{w} = \mathcal{D}(\mathbf{z}_0)$, where $\mathcal{D}: \mathbb{R}^{n \times d} \to \mathcal{V}^n$.

\section{Method}
\label{sec:method}

\begin{figure}
    \centering
    \begin{tabular}{c}
    \includegraphics[width=0.95\textwidth]{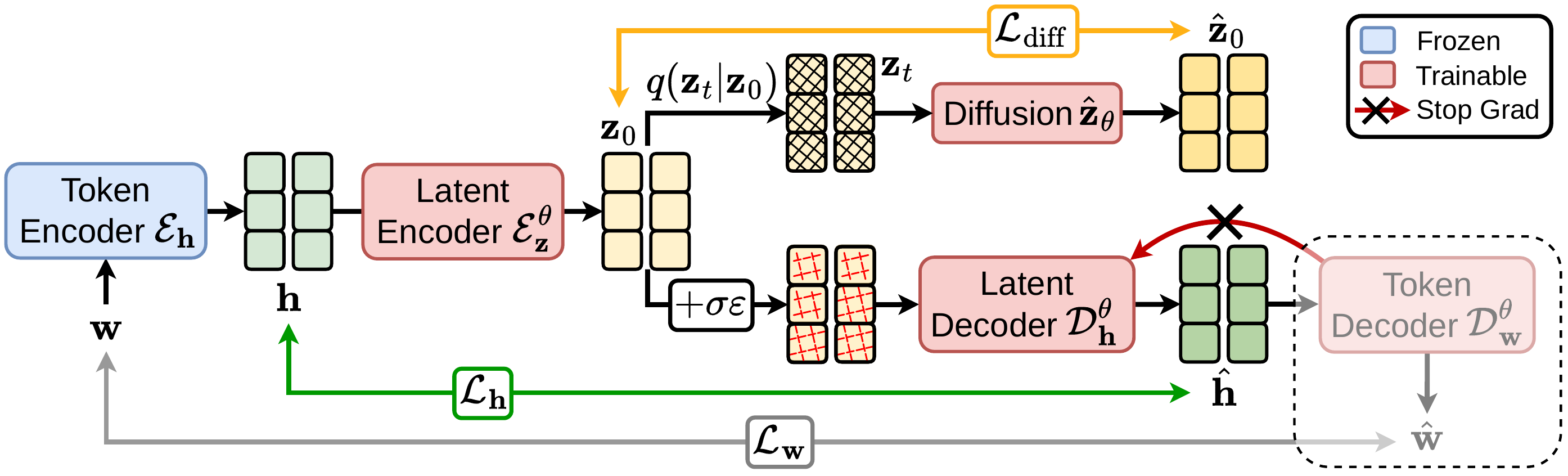}
    \end{tabular}
    \caption{
    \textbf{The proposed joint training framework.}
    Diffusion latents $\mathbf{z}_0$ are produced by applying a frozen pre-trained token encoder
    $\mathcal{E}_{\mathbf{h}}$ followed by a trainable latent encoder
    $\mathcal{E}_{\mathbf{z}}^{\theta}$ to the input tokens $\mathbf{w}$.
    The latents are decoded back to text through a trainable latent decoder
    $\mathcal{D}_{\mathbf{h}}^{\theta}$ and token decoder
    $\mathcal{D}_{\mathbf{w}}^{\theta}$.
  }
  \label{fig:framework}
\end{figure}

We propose \textbf{LDLM}, a latent text diffusion framework (\cref{fig:framework}) in which the latent space is learned jointly with a diffusion model, rather than fixed in advance by a separately trained autoencoder. In this section, we describe the design of the framework and the training objectives.

\subsection{Latent autoencoder}

\paragraph{Two-stage encoding.}
We encode text in two stages. 
Given an input token sequence $\mathbf{w}$, we first apply a frozen pre-trained token encoder $\mathcal{E}_{\mathbf{h}}$ (GPT-2 \citep{gpt2}) and extract contextual hidden states
$\mathbf{h} = \mathcal{E}_{\mathbf{h}}(\mathbf{w}).$
We then map these hidden states to the diffusion latent space with a trainable latent encoder,
$\mathbf{z}_0 = \mathcal{E}_{\mathbf{z}}^{\theta}(\mathbf{h}).$
The token encoder provides a contextual representation of the sequence, while the latent encoder reshapes this representation so that it remains decodable and is better suited for diffusion.

\paragraph{Two-stage decoding.}
We also decode latents in two stages. 
During training, we perturb the input of the latent decoder with Gaussian noise, $\tilde{\mathbf{z}}_0 = \mathbf{z}_0 + \sigma_{\mathrm{dec}} \epsilon$.
The role of this corruption is discussed in \cref{sec:decoder_noise}. 
The perturbed latent is first mapped to the token-encoder hidden state with the latent decoder, $\hat{\mathbf{h}} = \mathcal{D}_{\mathbf{h}}^\theta (\tilde{\mathbf{z}}_0)$, which is then converted into tokens with the token decoder, $\mathbf{w} = \mathcal{D}_{\mathbf{w}}^{\theta}(\hat{\mathbf{h}})$.
We train a latent decoder to align its predictions with the source hidden states:
\begin{equation}
\label{eq:h_loss}
\mathcal{L}_{\mathbf{h}}(\theta)
=
\mathbb{E}_{\mathbf{h},\,\epsilon\sim\mathcal{N}(\mathbf{0},\mathbf{I})}
\left[
\left\|
\mathbf{h}
-
\mathcal{D}_{\mathbf{h}}^{\theta}
\left(
\mathcal{E}_{\mathbf{z}}^{\theta}(\mathbf{h})
+
\sigma_{\mathrm{dec}}{\epsilon}
\right)
\right\|_2^2
\right].
\end{equation}
We use this loss as the only reconstruction signal that affects the latent encoder. When training the token decoder, we stop its gradient at the reconstructed hidden states to \emph{avoid affecting other models}.
\begin{equation}
\label{eq:w_loss}
\mathcal{L}_{\mathbf{w}}(\theta)
=
-
\mathbb{E}_{\mathbf{w},\,\epsilon\sim\mathcal{N}(\mathbf{0},\mathbf{I})}
\Big[
\log p_{\mathcal{D}_{\mathbf{w}}^{\theta}}
\big(
\mathbf{w} | \operatorname{sg}(\hat{\mathbf{h}})
\big)
\Big].
\end{equation}
where $\operatorname{sg}(\cdot)$ denotes the stop-gradient operator. It is important to note that to speed up training, the token decoder can be discarded during diffusion training and tuned only after other models have converged without any loss in the final quality. We empirically show this in Appendix \ref{app:decoder_post_training}. We further discuss the impact of each loss and the motivation for the proposed design in \cref{sec:recipe,sec:ablations}.

\subsection{Diffusion model}

We model the latent representations $\mathbf{z}_0 = \mathcal{E}_{\mathbf{z}}^{\theta}(\mathcal{E}_{\mathbf{h}}(\mathbf{w}))$ with a continuous diffusion model. 
Following \cref{sec:preliminaries}, we apply the forward noising process in the latent space and train the denoiser to recover the clean latent $\mathbf{z}_0$ from its noisy version $\mathbf{z}_t$ using the objective $\mathcal{L}_{\mathrm{diff}}$ from Eq \ref{eq:diff_loss}.
Following prior work~\citep{ld4lg, cosmos}, we also use self-conditioning~\citep{analog_bits}, where the denoiser is optionally conditioned on a previous estimate of $\mathbf{z}_0$.

Because the latents $\mathbf{z}_0$ are produced by the trainable latent encoder, the diffusion objective directly affects the latent space.
This is the key difference from other latent diffusion pipelines in which the encoder is frozen during diffusion training.
In practice, we control the early strength of this diffusion-to-encoder signal with the warmup described in the next section.

\subsection{Overall objective and sampling}\label{sec:loss_sampling}

The overall training objective admits an ELBO-style interpretation. We provide its careful derivation in Appendix~\ref{app:elbo}. 
In our implementation, we optimize
\begin{equation}
\label{eq:overall_loss}
\mathcal{L}(\theta) =
\mathcal{L}_{\mathrm{diff}}(\theta) +
\mathcal{L}_{\mathbf{h}}(\theta) +
\mathcal{L}_{\mathbf{w}}(\theta).
\end{equation}
At inference time, we sample Gaussian noise $\mathbf{z}_1 \sim \mathcal{N}(\mathbf{0},\mathbf{I})$ and run the reverse diffusion process with the Euler-Maruyama solver to obtain a clean latent prediction $\hat{\mathbf{z}}_0$. 
We then decode this latent without decoder-input corruption:
$\hat{\mathbf{h}}=\mathcal{D}_{\mathbf{h}}^{\theta}(\hat{\mathbf{z}}_0)$, and
$\hat{\mathbf{w}}\sim \mathcal{D}_{\mathbf{w}}^{\theta}(\hat{\mathbf{h}})$.
Thus, decoder-input noise is used only during training, while sampling always decodes the clean generated latent.

\section{Joint training recipe}
\label{sec:recipe}

In this section, we discuss the proposed design choices for the joint model training, which make up a four-part recipe. This includes the use of the pre-trained token encoder and the mean-square error (MSE) loss for the latent decoder, diffusion-to-encoder warmup, adaptive timestep sampling and the latent decoder input noise $\sigma_\mathrm{dec}\epsilon$.

\subsection{Pre-trained encoder and hidden state reconstruction}
\label{sec:pre-trained-representations}

A central design choice in LDLM is to construct the latent space on top of representations $\mathbf{h}$ produced by a frozen pre-trained token encoder $\mathcal{E}_\mathbf{h}$. 
Empirically, we found that learning a diffusion-friendly latent space is much easier when the trainable latent encoder $\mathcal{E}_\mathbf{z}^\theta$ operates on contextual hidden states, rather than directly on shallow token-level features.
These hidden states already encode dependencies across the sequence and provide a favorable starting point for latent learning.
This observation is consistent with recent work in image generation~\citep{zheng2025diffusion, chen2025aligning, kouzelis2025boosting}, where pre-trained DINO features provide a strong representation space and supervision signal for diffusion models, as well as with prior work on latent language diffusion~\citep{cosmos, ld4lg, ladd}. 

The frozen token encoder also allows us to incorporate the MSE loss $\mathcal{L}_{\mathbf{h}}$ between the hidden states $\mathbf{h}$ and outputs of the latent decoder $\mathcal{D}_\mathbf{h}^\theta$, instead of relying only on the cross-entropy (CE) loss as done in prior work \citep{diffusion-lm, diffuseq, seqdiffuseq}. In fact, this loss has a very important distinction from the CE loss. Both losses prevent the latent space collapse. However, as we add noise to the decoder input during training, CE loss forces all latents $\mathbf{z}_0$ to be well-separated for the decoder to accurately reconstruct the source tokens. At the same time, MSE loss is not that strict. If two latents are very close in the latent space, most probably their hidden states are also close. It means that it should be sufficient for the decoder to output an average hidden state of these two latents to get a low MSE loss. Therefore, \textbf{MSE loss does not strongly force latent separation}, but still prevents the latent space collapse. We find this distinction crucial for the diffusion model, and in \cref{sec:ablations}, we empirically show that MSE loss is essential for learning a robust latent space.

\subsection{Diffusion-to-encoder warmup}
\label{sec:diffusion-to-encoder-warmup}

At the beginning of joint training, the reconstruction $\mathcal{L}_\mathbf{h}$ and the diffusion $\mathcal{L}_{\mathrm{diff}}$ objectives pull the latent space in different directions: the reconstruction loss encourages the encoder to increase latent magnitudes to simplify decoding, whereas the diffusion loss encourages latents to shrink to make the denoising task trivial.
Empirically we found that the encoder struggles with learning meaningful representations, when both objectives affect it from the start of the training.

To avoid this early failure mode, we warm up the encoder by training it only with the reconstruction loss for several iterations and then gradually introduce the diffusion objective. This provides some time for the encoder to construct a decodable latent space before the diffusion objective starts shaping it.
The diffusion model itself is trained from the beginning, but gradients from $\mathcal{L}_{\mathrm{diff}}$ to the latent encoder are multiplied by a coefficient $\gamma(s)$ that increases from $\gamma_{\min} \approx 0$ to $1$. 
Details of the gradient scaling and the schedule for $\gamma(s)$ are given in Appendix \ref{app:diffusion-to-encoder-warmup-details}, and the warmup schedule is visualized in \cref{fig:warmup_schedule}. We study the effect of the warmup in \cref{sec:ablations}.

\subsection{Adaptive timestep sampling}
\label{sec:adaptive-time-sampling}

Prior work has shown the noise schedule has a substantial impact on diffusion model quality \citep{chen2023importancenoiseschedulingdiffusion} and its optimal form largely depends on latent-space properties \citep{difformer, tencdm, lee2026flow}. Therefore, it is important to select an optimal one. However, in our setting, the latent space evolves throughout training, which makes a fixed schedule challenging to tune. We therefore adapt the noise schedule dynamically during training, following the approach of \citep{cdcd}. 


We aim for the denoising loss to grow linearly with the sampled timestep, so that all timesteps contribute equally to reducing uncertainty.
Concretely, denoting by $\mathcal{L}(u)$ the expected loss at timestep $u \in [0, 1]$, we seek a sampling scheme such that, after a reparameterization $u = F^{-1}(t)$ with $t \sim \mathcal{U}[0, 1]$, $\mathcal{L}(u) - \mathcal{L}(0) = at$ for some constant $a > 0$. Since $\mathcal{L}$ is monotonically increasing in $u$ (as denoising becomes progressively harder), we set $F$ to be a cumulative distribution function (CDF) with density $p(u) \propto \frac{d\mathcal{L}}{du}$. Then, $t = F(u) = \int_0^u p(s)\, ds \propto \mathcal{L}(u) - \mathcal{L}(0) = at$. Therefore, sampling timesteps $u$ from a density proportional to $d\mathcal{L}/du$ produces a loss curve that is linear in $t$.

The only problem is that the derivative $d\mathcal{L}/du$ is not available in closed form. Thus, following \citep{cdcd, durkan2019neuralsplineflows}, we approximate it by partitioning $[0, 1]$ into $N = 100$ equal intervals via bin edges $0 = u_0 < u_1 < \dots < u_N = 1$, and tracking an exponential moving average estimate of $\mathcal{L}(u_i)$ at each edge throughout training. In this setting, $p(u) \propto \frac{d\mathcal{L}}{du}$ translates to its discrete counterpart 
\begin{equation}
p_i = \Pr\!\left[u \in (u_i, u_{i+1}]\right] = \frac{\mathcal{L}(u_{i+1}) - \mathcal{L}(u_i)}{ \mathcal{L}(u_N) - \mathcal{L}(u_0)}.
\end{equation}
At each training step, we first draw a bin index $i$ according to $\{p_i\}$ and then sample $u \sim \mathcal{U}(u_i, u_{i+1}]$. We update probabilities $p_i$ every $5{,}000$ iterations.



\subsection{Decoder noise $\sigma_{\mathrm{dec}}$}
\label{sec:decoder_noise}

We inject Gaussian noise with standard deviation $\sigma_{\mathrm{dec}}$ into the input of the latent decoder $\mathcal{D}_{\mathbf{h}}^\theta$ (see Eq.~\ref{eq:h_loss}). Although this might seem redundant, this noise plays three important roles. 
(1) When training the autoencoder, the dimensionality of the latent space often exceeds the dimensionality of its underlying manifold~\citep{mu2018allbutthetop}. The unused dimensions tend to contain random variation, which consumes the latent capacity and corrupts the training signal for the diffusion model \cite{dieleman2025latents}. By adding noise to the decoder input, we make reconstruction harder and encourage the encoder to \textbf{store useful information more efficiently} in order to preserve high reconstruction accuracy while keeping the latent norm low.
(2) The injected noise acts as data augmentation for the decoder, making it more \textbf{robust to errors} from the diffusion model \citep{tencdm}.
(3) Like any other model, diffusion models perform best when their inputs have the same variance \citep{edm}. However, without injected noise, the variance of latent coordinates collapses toward zero, causing the input variance to vary with $t$. Adding noise forces this variance to grow, which \textbf{improves the diffusion parameterization}.
We treat $\sigma_{\mathrm{dec}}$ as a hyperparameter and empirically study its effect in ablations (Section~\ref{sec:ablations}).

\section{Ablation study}
\label{sec:ablations}

In this section, we ablate the proposed joint training recipe and analyze the roles of its components.


\paragraph{Experimental setup.}
We use a GPT-2 model as a pre-trained token encoder $\mathcal{E}_\mathbf{h}$ and extract hidden representations $\mathbf{h}$ from the third to last layer. All representations are normalized to have a zero mean and a unit variance with dataset statistics. Both latent encoder $\mathcal{E}^\theta_\mathbf{z}$ and decoder $\mathcal{D}^\theta_\mathbf{h}$ employ 6-layer Perceiver Resampler \citep{flamingo} architecture. We use 12-layer Diffusion Transformer (DiT) \citep{diffusion_transformer} architecture for the diffusion model.
We set the default diffusion-to-encoder warmup length $S_{\mathbf{wu}} = 10$k, $\sigma_{\mathrm{dec}} = 1$ and employ adaptive timestep sampler.
We ablate the architectural choices in Appendix~\ref{app:architecture_ablations}.

We run the experiments on the OpenWebText (OWT)~\citep{Gokaslan2019OpenWeb} dataset, cropped to $128$ tokens to speed up training. We process all texts with GPT-2 tokenizer \citep{gpt2} and apply sequence packing during training.

\paragraph{Metrics.}
Following previous works, we utilize several metrics for quality evaluation. We measure the sample quality using the \textbf{Gen. PPL} computed with GPT-2 Large \citep{gpt2}. Knowing that Gen. PPL tends to be low for repetitive texts \citep{holtzman2020the}, we also estimate token diversity with two metrics: the n-gram \textbf{diversity}~\citep{su2022a} that measures the corpus-level diversity, and the unigram \textbf{entropy}~\citep{cdcd} that measures the sample-level diversity. We discuss the difference between them more thoroughly in Appendix~\ref{app:metrics}. In addition, we report \textbf{Mauve}~\citep{mauve}, which directly measures the proximity between the distributions of generated and reference texts.
All metrics are computed over five random seeds, each using $1{,}000$ generated texts.
All metric values are provided in a $\mathrm{mean}_{\pm \mathrm{std}}$ notation.

\begin{figure}
    \centering
    \begin{tabular}{c}
    \includegraphics[width=0.285\textwidth]{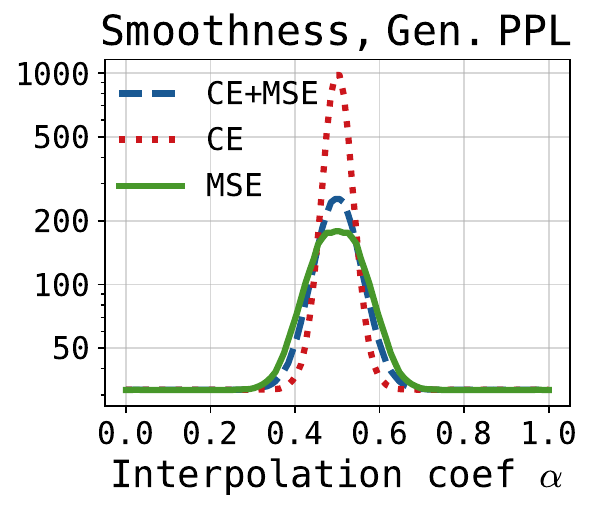}
    \includegraphics[width=0.28\textwidth]{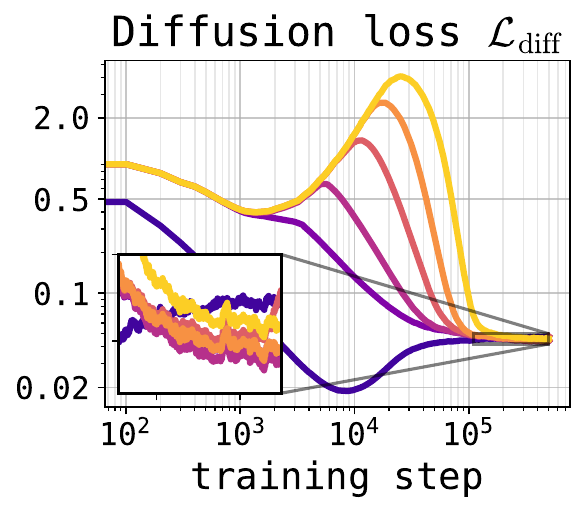}
    \includegraphics[width=0.37\textwidth]{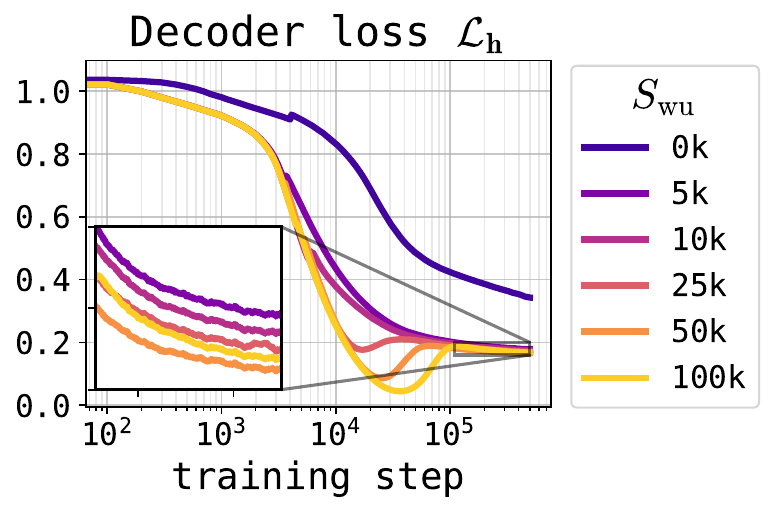}
    \end{tabular}
    \caption{\textbf{(Left):} Latent space smoothness for different decoder losses measured by Gen. PPL of decoded interpolated latents. \textbf{(Middle and Right):} Training loss dynamics for varying diffusion-to-encoder warmup size $S_{\mathrm{wu}}$.}
  \label{fig:ablations}
\end{figure}

\paragraph{Reconstruction loss.}
We first study how the choice of reconstruction objective affects the learned latent space.
We compare three variants: token-level cross-entropy $\mathcal{L}_\mathbf{w}$, hidden-state MSE $\mathcal{L}_\mathbf{h}$, and their combination. We turn off the stop-gradient in CE loss $\mathcal{L}_\mathbf{w}$ for this ablation to evaluate its impact on the latent space.
\cref{tab:ablation} shows that hidden-state MSE yields substantially better generation quality than token-level cross-entropy. 
Moreover, adding CE to the MSE objective degrades performance, suggesting that token-level supervision conflicts with the smooth latent geometry needed for diffusion.
To test this hypothesis more directly, we measure latent-space smoothness following prior work on latent diffusion for text~\citep{planner, cosmos}: we interpolate between two random latent representations of real texts, decode the interpolated points, and compute Gen. PPL along the interpolation path.
As shown in \cref{fig:ablations} (left), hidden-state MSE produces a smoother interpolation trajectory than token-level supervision, supporting its use as the reconstruction signal that shapes the latent space.

\begin{wraptable}{r}{0.48\textwidth}
\vspace{-0.2em}
\caption{Ablation study on OWT-128 dataset.}
\label{tab:ablation}
\small
\centering
\setlength{\tabcolsep}{2.7pt}
\begin{tabular}{lccc}
\toprule
Configuration & Gen. PPL ($\downarrow$) & Mauve ($\uparrow$) & Ent. ($\uparrow$) \\
\midrule
Real texts & $25.7_{\pm 0.4}$ & $100.0$ & $4.27_{\pm 0.00}$ \\
\midrule
\multicolumn{4}{l}{\textit{Reconstruction loss}} \\
\;\;CE & $156.3_{\pm 2.6}$ & $15.1_{\pm 1.4}$ & $4.21_{\pm 0.01}$ \\
\rowcolor{gray!20} \;\;\textbf{MSE} & $\mathbf{98.5}_{\pm 1.9}$ & $\mathbf{81.7}_{\pm 3.5}$ & $4.23_{\pm 0.01}$ \\
\;\;CE + MSE & $124.8_{\pm 2.4}$ & $64.9_{\pm 5.6}$ & $4.25_{\pm 0.01}$\\
\midrule
\multicolumn{4}{l}{\textit{Token encoder (CE loss)}} \\
\rowcolor{gray!20}\;\;$\boldsymbol{\mathcal{E}_\mathbf{h}(\cdot)}$ & $\mathbf{156.3}_{\pm 2.6}$ & $\mathbf{15.1}_{\pm 1.4}$ & $4.21_{\pm 0.01}$ \\
\;\;\textsc{Emb}$(\cdot)$ & $190.4_{\pm 3.4}$ & $3.9_{\pm 0.8}$ & $4.18_{\pm 0.00}$ \\
\;\;\textsc{Emb}$_\theta(\cdot)$ & $272.4_{\pm 8.0}$ & $3.0_{\pm 0.4}$ & $4.24_{\pm 0.01}$ \\
\midrule
\multicolumn{4}{l}{\textit{Diffusion-to-encoder warmup $S_{\mathbf{wu}}$ (w/o ATS)}} \\
\;\;No warmup & $403.1_{\pm 10.8}$ & $0.4_{\pm 0.1}$ & $3.11_{\pm 0.01}$ \\
\;\;5\,000 & $274.7_{\pm 6.9}$ & $4.3_{\pm 0.6}$ & $4.28_{\pm 0.00}$ \\
\;\;10\,000 & $171.9_{\pm 2.2}$ & $11.8_{\pm 1.9}$ & $4.24_{\pm 0.00}$ \\
\;\;25\,000 & $125.3_{\pm 3.1}$ & $53.6_{\pm 2.5}$ & $4.25_{\pm 0.01}$ \\
\rowcolor{gray!20}\;\;\textbf{50\,000} & $\mathbf{116.7}_{\pm 3.3}$ & $67.0_{\pm 3.2}$ & $4.24_{\pm 0.01}$ \\
\;\;100\,000 & $119.4_{\pm 2.5}$ & $68.8_{\pm 2.7}$ & $4.25_{\pm 0.01}$ \\
\midrule
\multicolumn{4}{l}{\textit{Timestep sampling}} \\
\;\;Uniform & $171.9_{\pm 2.2}$ & $11.8_{\pm 1.9}$ & $4.24_{\pm 0.01}$ \\
\rowcolor{gray!20}\;\;\textbf{Adaptive} & $\mathbf{98.5}_{\pm 1.9}$ & $\mathbf{81.7}_{\pm 3.5}$ & $4.23_{\pm 0.01}$ \\
\midrule
\multicolumn{4}{l}{\textit{Decoder input noise} $\sigma_\mathrm{dec}$} \\
\;\;$0.1$ & $702.2_{\pm 31.5}$ & $0.6_{\pm 0.1}$ & $3.58_{\pm 0.03}$ \\
\;\;$0.5$ & $288.9_{\pm 5.4}$ & $3.3_{\pm 0.2}$ & $4.23_{\pm 0.01}$ \\
\;\;$1.0$ & $98.5_{\pm 1.9}$ & $81.7_{\pm 3.5}$ & $4.23_{\pm 0.01}$ \\
\;\;$2.0$ & $74.2_{\pm 1.0}$ & $87.3_{\pm 0.7}$ & $4.24_{\pm 0.01}$ \\
\rowcolor{gray!20}\;\;$\mathbf{3.0}$ & $\mathbf{66.3}_{\pm 0.8}$ & $\mathbf{89.1}_{\pm 3.1}$ & $4.25_{\pm 0.01}$ \\
\;\;$4.0$ & $38.1_{\pm 0.6}$&$0.4_{\pm 0.1}$ & $2.19_{\pm 0.01}$ \\
\bottomrule
\end{tabular}
\vspace{-2em}
\end{wraptable}

\paragraph{Token encoder.}
Next, we study the impact of the token encoder by varying its architecture. We compare three different variants: GPT-2 model, shallow token embeddings from GPT-2, and trainable token embeddings learned from scratch. As it is impossible to use the MSE loss $\mathcal{L}_\mathbf{h}$ for the latter setup, we use CE reconstruction objective $\mathcal{L}_\mathbf{w}$ without stop-gradient for all encoders to ensure fair comparison.
\cref{tab:ablation} shows that contextual hidden states substantially outperform both embedding-based alternatives.
This suggests that the latent encoder benefits not merely from a continuous input space, but from contextual features that capture sequence-level structure.

\paragraph{Diffusion-to-encoder warmup.}
In order to study the effect of diffusion-to-encoder warmup, we
compare warmup lengths $S_{\mathrm{wu}}\in\{0, 5\mathrm{k}, 10\mathrm{k}, 25\mathrm{k}, 50\mathrm{k}, 100\mathrm{k}\}$, disabling adaptive timestep sampling and keeping all other components fixed.
Without warmup, the diffusion objective dominates early updates to the latent encoder and quickly reduces the latent magnitude, which is reflected in rapidly decreasing diffusion loss in \cref{fig:ablations} (middle).
This latent norm constraint prevents the token encoder from learning meaningful features, which hurts the decoder performance, as shown in \cref{fig:ablations} (right).
Thus, generation quality degrades, with high generative perplexity and low entropy in \cref{tab:ablation}.

As described in \cref{sec:diffusion-to-encoder-warmup}, warmup removes all constraints from the early stages of encoder training, allowing the encoder to learn text semantics before introducing the diffusion objective. In \cref{fig:ablations} (right) and \cref{tab:ablation}, we observe that sufficient warmup ($50$k$+$ steps) improves decoder quality, which in turn benefits overall generation.
The exact warmup length is not critical once it is long enough to form a decodable latent space, but overly long warmup delays diffusion-driven shaping of the latent space, leaving fewer training iterations for convergence.

\begin{wrapfigure}{r}{0.3\textwidth}
    \vspace{-0.3em}
    \centering
    \includegraphics[width=0.3\textwidth]{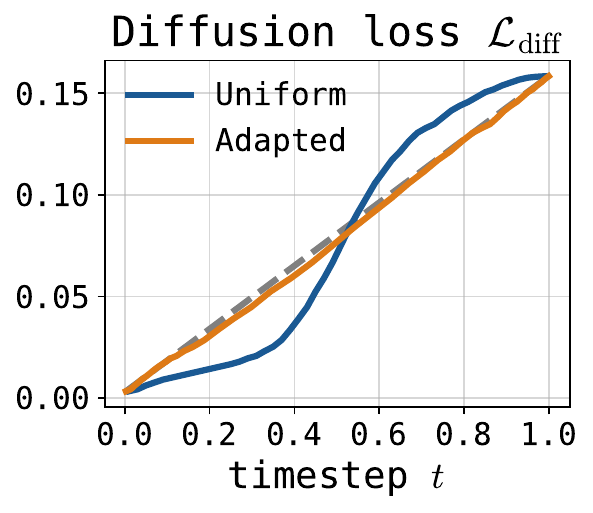}
    \vspace{-1em}
    \caption{Diffusion loss w.r.t timestep for the uniform time $t$ and adapted time $u = F^{-1}(t)$.}
    \label{fig:ast}
    \vspace{-1.5em}
\end{wrapfigure}

\paragraph{Adaptive timestep sampling.}
We compare uniform timestep sampling with the adaptive sampler, keeping all other components fixed and using $S_{\mathrm{wu}}=10\mathrm{k}$.
As shown in \cref{tab:ablation}, replacing uniform sampling with adaptive sampling substantially improves generative perplexity and Mauve while keeping entropy essentially unchanged.
\cref{fig:ast} confirms that the adaptive sampler makes the diffusion loss growth linear with respect to the timestep $t$, while the fixed noise schedule produces notably worse distribution of denoising difficulty.


\begin{figure}
    \hspace*{-0.4cm}
    \centering
    \begin{tabular}{c}
    \includegraphics[width=0.355\textwidth]{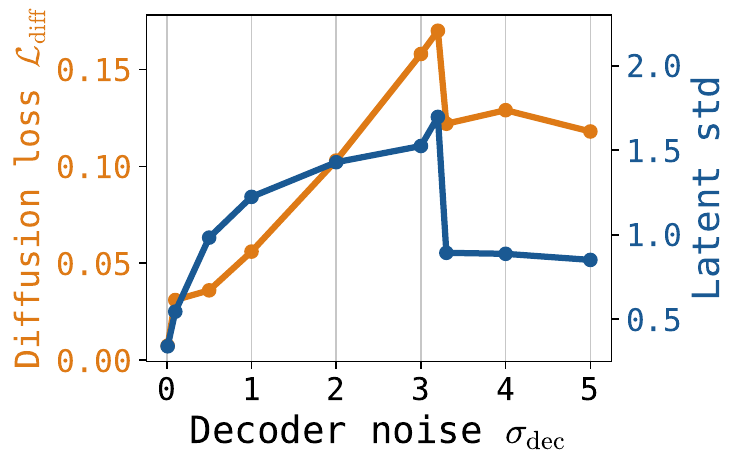}
    \includegraphics[width=0.35\textwidth]{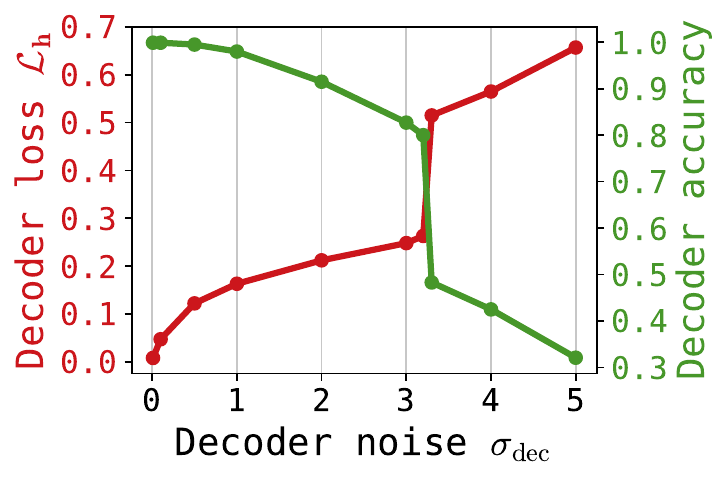}
    \includegraphics[width=0.29\textwidth]{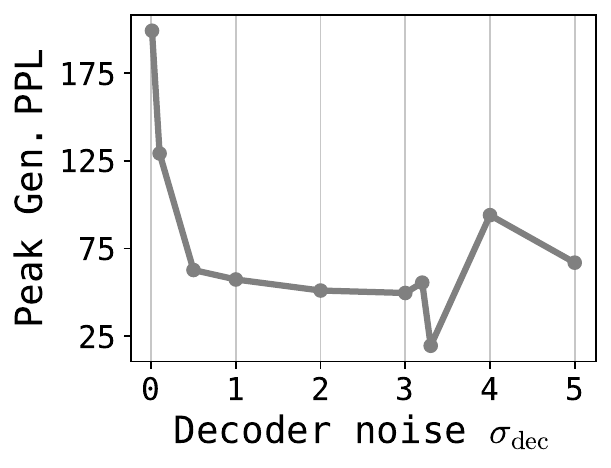}
    \end{tabular}
    \vspace{-0.4em}
    \caption{
    Effect of decoder noise $\sigma_{\mathrm{dec}}$ on the model quality.
    \textbf{(Left):} Latent space state reflected by the diffusion loss and coordinate-wise latent standard deviation. \textbf{(Middle):} Decoder loss and reconstruction accuracy. \textbf{(Right):} Latent space smoothness measured as Gen. PPL of the decoded mean of two random latents.
    }
  \label{fig:decoder_sigma}
  \vspace{-1em}
\end{figure}

\paragraph{Decoder input noise.}
Finally, we study the effect of Gaussian noise injected into the latent decoder input.
We compare several values of $\sigma_{\mathrm{dec}}$ and provide the training results in 
\cref{tab:ablation} and \cref{fig:decoder_sigma}. We observe that decoder-input noise has a non-monotonic effect.
For small values of $\sigma_{\mathrm{dec}}$, reconstruction remains easy, interpolation smoothness is poor, and diffusion quality remains low.
Increasing $\sigma_{\mathrm{dec}}$ improves generation quality by making the latent space smoother and the latent decoder more robust to diffusion errors.

This improvement holds only up to a finite noise level.
When $\sigma_{\mathrm{dec}}$ becomes too large, the system starts breaking: the latent space tries to expand to keep the reconstruction loss low, but the diffusion loss outweighs this expansion, as MSE scales quadratically with the latent norm. The decoder becomes unable to accurately reconstruct $\mathbf{h}$, which leads to poor token reconstruction and, therefore, significant drop in text quality.
The results at $\sigma_{\mathrm{dec}}=4.0$ illustrate this failure mode: generative perplexity becomes low, but Mauve and entropy drop sharply, indicating collapsed generations rather than better text quality.
We therefore use an intermediate value, $\sigma_{\mathrm{dec}}=3$ in all further experiments to balance latent regularization, decoder robustness, and reconstruction accuracy.

\section{Latent space comparison}
\label{sec:space_comparison}

\begin{wraptable}{r}{0.48\textwidth}
\vspace{-1.0em}
\caption{Comparison of latent spaces for diffusion model training on OWT-128.}
\label{tab:space_comparison}
\small
\centering
\setlength{\tabcolsep}{2.7pt}
\begin{tabular}{lccc}
\toprule
Space & Gen. PPL ($\downarrow$) & Mauve ($\uparrow$) & Ent. ($\uparrow$) \\
\midrule
Scratch emb & $570.1_{\pm 15.4}$ & $0.8_{\pm 0.1}$ & $4.26_{\pm 0.01}$ \\
GPT-2 emb & $299.2_{\pm 3.4}$ & $1.6_{\pm 0.2}$ & $4.26_{\pm 0.00}$ \\
TEncDM & $149.7_{\pm 1.8}$ & $36.6_{\pm 5.9}$ & $4.28_{\pm 0.00}$ \\
Frozen enc. & $152.3_{\pm 1.5}$ & $34.0_{\pm 5.3}$ & $4.25_{\pm 0.01}$ \\
COSMOS & $98.5_{\pm 1.8}$ & $53.4_{\pm 4.8}$ & $4.19_{\pm 0.01}$ \\
\rowcolor{gray!20} \textbf{LDLM} & $\mathbf{66.3}_{\pm 0.8}$ & $\mathbf{89.1}_{\pm 3.1}$ & $4.25_{\pm 0.01}$ \\
\bottomrule
\end{tabular}
\end{wraptable}

In this section, we compare LDLM with the diffusion latent spaces employed in prior work, demonstrating that the most ``diffusable'' latent space can only be obtained when it is tuned jointly with the diffusion model. Our comparison covers progressively more sophisticated approaches to latent space construction: shallow token embeddings trained from scratch with diffusion \citep{diffusion-lm, cdcd}, frozen token embeddings \citep{sed}, hidden states of a pre-trained encoder (TEncDM-like \citep{tencdm}), a pretrained and frozen latent encoder (Frozen enc.), and latents produced by an autoencoder trained with explicit smoothing and robustness objectives (COSMOS-like \citep{cosmos}). For the Frozen enc. baseline, we train the same latent autoencoder as LDLM with the same reconstruction objective and $\sigma_{\mathrm{dec}}=3$ for 500k steps, then freeze the latent encoder and train diffusion for another 500k steps. Across all setups, we use GPT-2 as the encoder backbone and the GPT-2 tokenizer. We also use the same diffusion and decoder architectures to ensure that the latent space is the only varying factor.

Table~\ref{tab:space_comparison} reports the comparison results on OWT-128. Shallow token embeddings perform poorly, even when initialized from GPT-2. Switching to contextual GPT-2 embeddings substantially improves generation quality, confirming the importance of sequence-level information for diffusion. The Frozen enc. baseline performs on par with fixed GPT-2 hidden states, while explicit smoothing improves generation quality. Nevertheless, jointly trained LDLM latents achieve the best results among all methods and are also the smoothest in an interpolation test (Appendix~\ref{app:latent_smoothness}). LDLM also outperforms TEncDM and COSMOS-style baselines across all four tested GPT-2 layers (Appendix~\ref{app:gpt_layers}). Replacing GPT-2 with bidirectional BERT-base preserves strong generation quality without additional hyperparameter tuning (Appendix~\ref{app:pretrained_encoder}).

These findings indicate that diffusion benefits when high-level sequence information is already encoded in the latents. Intuitively, reconstructing a noisy sequence requires first understanding its global content and then recovering its clean version. A sufficiently expressive encoder produces latents in which contextual information has already been extracted, simplifying the denoising task. LDLM pushes this logic further. Instead of relying on a latent space optimized for another objective and hoping that it is suitable for diffusion, we train the encoder jointly with the diffusion model. This gives us direct control over how information is organized in the latent space, allowing it to be shaped specifically for the diffusion model.

\section{Main results}
\label{sec:main_results}

We evaluate LDLM on unconditional generation with LM1B~\citep{chelba2013one} (length 128) and OpenWebText (OWT)~\citep{Gokaslan2019OpenWeb} (length 1024), using the setup described in \cref{sec:ablations} with $S_\mathrm{wu} = 50$k and $\sigma_\mathrm{dec} = 3$. Other implementation details are provided in Appendix~\ref{app:impl_det}.
We compare against representative diffusion-based text generation methods: masked diffusion MDLM~\citep{mdlm} with its ReMDM inference variant~\citep{wang2025remasking}, uniform discrete diffusion Duo~\citep{duo} with the $\Psi$-sampler~\citep{duo_ch2}, hybrid discrete-continuous diffusion CANDI~\citep{candi}, and one-hot flow matching FLM~\citep{lee2026flow}.
We use MDLM and ReMDM as LLaDA-style masked diffusion baselines at a comparable scale on our benchmarks, since LLaDA~\citep{llada} uses a substantially larger model and training corpus.
We use publicly available checkpoints when possible; for CANDI, we evaluate a checkpoint provided by the authors.
For LM1B, we train the baselines using the official implementations.


\paragraph{Quality-diversity trade-off.} To capture both sample quality and diversity, we plot the Pareto frontiers for Gen. PPL and entropy obtained by varying the number of generation steps.
\cref{fig:owt_lm1b} summarizes the resulting quality-diversity trade-off; more detailed results are presented in Appendix~\ref{sec:full_comparison_tables}.
On both OWT and LM1B, LDLM yields the strongest Pareto frontier and comes close to the real-text quality. At comparable entropy, it achieves noticeably lower generative perplexity than the baselines.


\paragraph{Computational efficiency.}

\begin{wraptable}{r}{0.45\textwidth}
\vspace{-1.0em}
\centering
\caption{Training and sampling time on OWT-1024. Sampling uses 1024 NFEs.}
\label{tab:time}
\small
\setlength{\tabcolsep}{2.7pt}
\begin{tabular}{lcc}
\toprule
Method & Train (d) & Sample (s) \\
\midrule
MDLM~\citep{mdlm} & $2.9$ & $529$ \\
Duo~\citep{duo} & $5.5$ & $537$ \\
Duo$^{++}$ + $\Psi$-sampler~\citep{duo_ch2} & $3.2$ & $663$ \\
CANDI~\citep{candi} & $5.6$ & $142$ \\
FLM~\citep{lee2026flow} & $3.0$ & $97$ \\
\rowcolor{gray!20}
\textbf{LDLM} & $5.1$ & $\mathbf{40}$ \\
\bottomrule
\end{tabular}
\end{wraptable}

In addition to sampling quality, we compare the methods in terms of generation speed. The most prominent text diffusions operate with either discrete tokens \cite{mdlm, duo, wang2025remasking, duo_ch2} or one-hot token encodings \citep{lee2026flow, candi}. Both paradigms require projecting the denoiser outputs onto vocabulary-sized vectors at every denoising step, which substantially slows down generation. In contrast, LDLM acts in a lower-dimensional latent space throughout the entire generation process, leading to faster sampling.

We measure training and sampling times for LDLM and the competing baselines, reporting the results in Table~\ref{tab:time} and Figure~\ref{fig:owt_lm1b}, where sampling time is indicated by marker size. All measurements are performed on the OWT-1024 dataset with batch size $16$ and a single NVIDIA A100 GPU. We train each model for 1M iterations and use $1024$ NFEs for sampling. For LDLM, we discard the token decoder during training and tune it separately afterwards to avoid unnecessary overhead. While LDLM's training speed is limited by the need to invoke the autoencoder at every step, its sampling runs $2{\text -}13\times$ faster than all competing methods.

\section{Conclusion}
\label{sec:conclusion}
In this work, we introduced LDLM, a latent diffusion language model that trains the latent space jointly with the diffusion model.
We showed that naive joint training is not robust on its own, and proposed a simple recipe that makes this approach work.
On LM1B and OpenWebText, LDLM achieves a better quality-diversity trade-off while sampling faster than competing diffusion baselines.
These results suggest that shaping the latent space directly with the denoising objective is essential for building latent text diffusion models. We discuss the limitations of the work in Appendix~\ref{app:limitations}.

\begin{ack}
The work was supported by the grant for research centers in the field of AI provided by the Ministry of Economic Development of the Russian Federation in accordance with the agreement 000000C313925P4E0002 and the agreement with HSE University No.~139-15-2025-009. This research was supported in part through computational resources of HPC facilities at HSE University.
\end{ack}

\bibliography{bibliography}
\bibliographystyle{abbrv}

\newpage
\appendix
\onecolumn
\section*{Appendix}
\addtocontents{toc}{\protect\setcounter{tocdepth}{2}}
\addtocontents{toc}{\string\renewcommand{\string\cftsecfont}{\string\normalfont}}
\renewcommand{\cftsecleader}{\cftdotfill{\cftdotsep}}
\renewcommand{\contentsname}{} 

\cftsetindents{section}{0em}{2em}

\tableofcontents
\vspace{1em}  
\hrule

\newpage
\appendix
\newpage
\section{Additional method details}
\label{app:method_details}

\subsection{ELBO interpretation of the joint objective}
\label{app:elbo}

This appendix provides a probabilistic interpretation of the training objective used to jointly train
the latent encoder, the decoder, and the latent diffusion prior.
The key subtlety is that the encoder used in the main text is deterministic. A strictly deterministic posterior,
\begin{equation}
q_\theta(\mathbf z_0 \mid \mathbf w)
=
\delta\!\left(\mathbf z_0-\boldsymbol\mu_\theta(\mathbf w)\right),
\qquad
\boldsymbol\mu_\theta(\mathbf w)
=
E_z^\theta(E_h(\mathbf w)),
\end{equation}
does not admit a finite KL divergence to a continuous prior \(p_\theta(\mathbf z_0)\).
To obtain a well-defined ELBO, we therefore introduce the standard fixed-noise relaxation
\begin{equation}
q_\theta^{\tau}(\mathbf z_0 \mid \mathbf w)
=
\mathcal N\!\left(
\mathbf z_0;
\boldsymbol\mu_\theta(\mathbf w),
\tau^2 I
\right),
\label{eq:fixed_noise_posterior}
\end{equation}
where \(\tau > 0\) is fixed and not learned.

\paragraph{Observed variables.}
For an input token sequence \(\mathbf w\), the frozen language model produces hidden states
$\mathbf h = E_h(\mathbf w)$.
Since \(\mathbf h\) is a deterministic function of \(\mathbf w\), we can treat \((\mathbf w,\mathbf h)\) as an
augmented observation. This gives a direct probabilistic interpretation to the hidden-state reconstruction term.

\paragraph{Training-time latent corruption.}
The practical objective includes latent corruption in the reconstruction branch.
Namely, the decoder receives a Gaussian-perturbed latent
\(\tilde{\mathbf z}_0\) rather than the clean latent \(\mathbf z_0\), with a fixed noise variance.
We denote this decoder-input corruption kernel by
\begin{equation}
r(\tilde{\mathbf z}_0 \mid \mathbf z_0)
=
\mathcal N\!\left(
\tilde{\mathbf z}_0;
\mathbf z_0,
\sigma_{\mathrm{dec}}^2 I
\right),
\label{eq:decoder_corruption_kernel}
\end{equation}
where \(\sigma_{\mathrm{dec}}\) is the decoder-input noise used in Section~4.2.

\paragraph{Generative model.}
We model the latent prior with a diffusion model \(p_\theta(\mathbf z_0)\). Conditioned on the corrupted
latent \(\tilde{\mathbf z}_0\), the decoder first reconstructs hidden states and then predicts tokens.
We define the hidden-state likelihood as
\begin{equation}
p_\theta(\mathbf h \mid \tilde{\mathbf z}_0)
=
\mathcal N\!\left(
\mathbf h;
D_h^\theta(\tilde{\mathbf z}_0),
\sigma_{\mathrm{rec}}^2 I
\right),
\label{eq:hidden_likelihood}
\end{equation}
and define the token likelihood through the token decoder logits
\begin{equation}
\boldsymbol\ell(\tilde{\mathbf z}_0)
=
D_w^\theta\!\left(D_h^\theta(\tilde{\mathbf z}_0)\right),
\qquad
p_\theta(\mathbf w \mid \tilde{\mathbf z}_0)
=
\prod_{i=1}^n
\mathrm{Cat}\!\left(
w_i;\,
\mathrm{softmax}(\ell_i(\tilde{\mathbf z}_0))
\right).
\label{eq:token_likelihood}
\end{equation}
Thus, the joint model is
\begin{equation}
p_\theta(\mathbf w,\mathbf h,\tilde{\mathbf z}_0,\mathbf z_0)
=
p_\theta(\mathbf z_0)\,
r(\tilde{\mathbf z}_0 \mid \mathbf z_0)\,
p_\theta(\mathbf h \mid \tilde{\mathbf z}_0)\,
p_\theta(\mathbf w \mid \tilde{\mathbf z}_0).
\label{eq:joint_model}
\end{equation}

\paragraph{Variational posterior.}
We use the relaxed encoder posterior from ~\cref{eq:fixed_noise_posterior} together with the same
decoder corruption kernel:
\begin{equation}
q_\theta(\mathbf z_0,\tilde{\mathbf z}_0 \mid \mathbf w)
=
q_\theta^{\tau}(\mathbf z_0 \mid \mathbf w)\,
r(\tilde{\mathbf z}_0 \mid \mathbf z_0).
\label{eq:variational_posterior}
\end{equation}

\paragraph{ELBO derivation.}
Starting from the marginal likelihood of the augmented observation,
\begin{align}
\log p_\theta(\mathbf w,\mathbf h)
&=
\log
\int
p_\theta(\mathbf z_0)\,
r(\tilde{\mathbf z}_0 \mid \mathbf z_0)\,
p_\theta(\mathbf h \mid \tilde{\mathbf z}_0)\,
p_\theta(\mathbf w \mid \tilde{\mathbf z}_0)\,
d\tilde{\mathbf z}_0\,d\mathbf z_0
\nonumber\\
&=
\log
\mathbb E_{q_\theta(\mathbf z_0,\tilde{\mathbf z}_0\mid \mathbf w)}
\left[
\frac{
p_\theta(\mathbf z_0)\,
r(\tilde{\mathbf z}_0 \mid \mathbf z_0)\,
p_\theta(\mathbf h \mid \tilde{\mathbf z}_0)\,
p_\theta(\mathbf w \mid \tilde{\mathbf z}_0)
}{
q_\theta^{\tau}(\mathbf z_0 \mid \mathbf w)\,
r(\tilde{\mathbf z}_0 \mid \mathbf z_0)
}
\right].
\end{align}
The corruption terms \(r(\tilde{\mathbf z}_0 \mid \mathbf z_0)\) cancel, and we can write the ELBO as
\begin{equation}
\mathcal L_{\mathrm{ELBO}}
=
\mathbb E_{q_\theta^{\tau}(\mathbf z_0\mid \mathbf w)\,r(\tilde{\mathbf z}_0\mid \mathbf z_0)}
\Big[
\log p_\theta(\mathbf w \mid \tilde{\mathbf z}_0)
+
\log p_\theta(\mathbf h \mid \tilde{\mathbf z}_0)
\Big]
-
\mathrm{KL}
\left(
q_\theta^{\tau}(\mathbf z_0 \mid \mathbf w)
\,\|\, p_\theta(\mathbf z_0)
\right).
\label{eq:elbo}
\end{equation}

Equivalently, expanding the KL term gives
\begin{align}
\mathcal L_{\mathrm{NELBO}}
&=
\underbrace{
\mathbb E_{q_\theta^{\tau}(\mathbf z_0\mid \mathbf w)\,r(\tilde{\mathbf z}_0\mid \mathbf z_0)}
\left[
-\log p_\theta(\mathbf w \mid \tilde{\mathbf z}_0)
\right]
}_{\text{token reconstruction}}
+
\underbrace{
\mathbb E_{q_\theta^{\tau}(\mathbf z_0\mid \mathbf w)\,r(\tilde{\mathbf z}_0\mid \mathbf z_0)}
\left[
-\log p_\theta(\mathbf h \mid \tilde{\mathbf z}_0)
\right]
}_{\text{hidden-state reconstruction}}
\nonumber\\
&\hspace{2em}
+
\underbrace{
\mathbb E_{q_\theta^{\tau}(\mathbf z_0\mid \mathbf w)}
\left[
\log q_\theta^{\tau}(\mathbf z_0 \mid \mathbf w)
\right]
}_{\text{negative encoder entropy}}
+
\underbrace{
\mathbb E_{q_\theta^{\tau}(\mathbf z_0\mid \mathbf w)}
\left[
-\log p_\theta(\mathbf z_0)
\right]
}_{\text{latent prior cross-entropy}}.
\label{eq:expanded_elbo}
\end{align}

\paragraph{Encoder entropy is constant.}
Let \(d_z\) denote the total scalar dimensionality of \(\mathbf z_0\). Since the covariance
\(\tau^2 I\) in ~\cref{eq:fixed_noise_posterior} is fixed, we have
\begin{equation}
\mathbb E_{q_\theta^{\tau}(\mathbf z_0\mid \mathbf w)}
\left[
\log q_\theta^{\tau}(\mathbf z_0 \mid \mathbf w)
\right]
=
-\frac{d_z}{2}\log(2\pi e \tau^2),
\end{equation}
which is constant with respect to all trainable parameters.

\paragraph{Reconstruction terms.}
For reconstruction terms we obtain
\begin{equation}
-\log p_\theta(\mathbf h \mid \tilde{\mathbf z}_0)
=
\frac{1}{2\sigma_{\mathrm{rec}}^2}
\left\|
\mathbf h-D_h^\theta(\tilde{\mathbf z}_0)
\right\|_2^2
+
\mathrm{const}
.
\label{eq:hidden_nll}
\end{equation}

\begin{equation}
-\log p_\theta(\mathbf w \mid \tilde{\mathbf z}_0)
=
\mathrm{CE}\!\left(
\mathbf w,\,
D_w^\theta(D_h^\theta(\tilde{\mathbf z}_0))
\right)
.
\label{eq:token_nll}
\end{equation}


\paragraph{Diffusion bound for the latent prior term.}
The prior \(p_\theta(\mathbf z_0)\) is represented by a latent diffusion model.
Following \citep{vahdat2021score}, for a variance-preserving forward process
\begin{equation}
\mathbf z_t
=
\alpha_t \mathbf z_0+\sigma_t\epsilon_d,
\qquad
\epsilon_d\sim\mathcal N(0,I),
\end{equation}
with log-SNR
\begin{equation}
\lambda(t)
=
\log \frac{\alpha_t^2}{\sigma_t^2},
\end{equation}
we can upper-bound the latent prior cross-entropy as
\begin{equation}
-\log p_\theta(\mathbf z_0)
\le
C_{\mathrm{diff}}
+
\mathbb E_{t\sim p(t),\,\epsilon_d}
\left[
\omega_{\mathrm{ELBO}}(t)
\left\|
\mathbf z_0
-
\hat{\mathbf z}_\theta(\mathbf z_t,t)
\right\|_2^2
\right].
\label{eq:diffusion_elbo}
\end{equation}

For uniform \(t\), the usual likelihood-weighted coefficient in
\(\mathbf x_0\)-prediction form is
\begin{equation}
\omega_{\mathrm{ELBO}}(t)
=
-\frac{1}{2}\lambda'(t)\exp(\lambda(t)).
\end{equation}
The term \(C_{\mathrm{diff}}\) collects the terminal KL and schedule-dependent constants. For our diffusion schedule, this term degenerates into a constant because \(\alpha_t\) is exactly zero at the terminal time.

\paragraph{Final NELBO objective.}
Substituting all the expressions derived above into the negative ELBO, we obtain
\begin{align}
\mathcal L_{\mathrm{NELBO}}
&=
\mathbb E_{q_\theta^{\tau}(\mathbf z_0\mid \mathbf w)\,r(\tilde{\mathbf z}_0\mid \mathbf z_0)}
\left[
\mathrm{CE}\!\left(
\mathbf w,\,
D_w^\theta(D_h^\theta(\tilde{\mathbf z}_0))
\right)
\right]
\nonumber\\
&\hspace{2em}
+
\frac{1}{2\sigma_{\mathrm{rec}}^2}
\mathbb E_{q_\theta^{\tau}(\mathbf z_0\mid \mathbf w)\,r(\tilde{\mathbf z}_0\mid \mathbf z_0)}
\left[
\left\|
\mathbf h-
D_h^\theta(\tilde{\mathbf z}_0)
\right\|_2^2
\right]
\nonumber\\
&\hspace{2em}
+
\mathbb E_{q_\theta^{\tau}(\mathbf z_0\mid \mathbf w),\,t\sim p(t),\,\mathbf z_t\sim q(\mathbf z_t\mid \mathbf z_0)}
\left[
\omega_{\mathrm{ELBO}}(t)
\left\|
\mathbf z_0
-
\hat{\mathbf z}_\theta(\mathbf z_t,t)
\right\|_2^2
\right]
+
\mathrm{const},
\label{eq:final_elbo_style}
\end{align}

\paragraph{Connection to the implemented objective.}
The training loss used in the main text is motivated by the form of the NELBO above after
dropping constants, taking the deterministic-encoder limit \(\tau\to0\), and applying the
standard weighting simplifications: (i) the exact ELBO weighting \(\omega_{\mathrm{ELBO}}(t)\)
is replaced by the simple \(\mathbf x_0\)-prediction loss used in the main text, and (ii) the
factor \((2\sigma_{\mathrm{rec}}^2)^{-1}\) is absorbed into 1.
This gives
\begin{equation}
\mathcal L
=
\mathcal L_{\mathrm{diff}}
+
\mathcal L_h
+
\mathcal L_w,
\end{equation}
with
\begin{align}
\mathcal L_{\mathrm{diff}}
&=
\mathbb E_{t,\epsilon_d}
\left[
\left\|
\mathbf z_0
-
\hat{\mathbf z}_\theta(\mathbf z_t,t)
\right\|_2^2
\right],
\\
\mathcal L_h
&=
\mathbb E_{\epsilon_{\mathrm{dec}}}
\left[
\left\|
\mathbf h-
D_h^\theta(\mathbf z_0+\sigma_{\mathrm{dec}}\epsilon_{\mathrm{dec}})
\right\|_2^2
\right],
\\
\mathcal L_w
&=
\mathbb E_{\epsilon_{\mathrm{dec}}}
\left[
\mathrm{CE}\!\left(
\mathbf w,\,
D_w^\theta(D_h^\theta(\mathbf z_0+\sigma_{\mathrm{dec}}\epsilon_{\mathrm{dec}}))
\right)
\right].
\end{align}
\paragraph{Training-time corruption versus inference.}
The corruption kernel \(r(\tilde{\mathbf z}_0 | \mathbf z_0)\) is introduced to interpret the
noise-augmented decoder losses used during training.
At inference time, however, the sampling procedure described in \cref{sec:loss_sampling} decodes the clean generated latent
\(\hat{\mathbf z}_0\) directly without any noise injection.

\subsection{Diffusion-to-encoder warmup}
\label{app:diffusion-to-encoder-warmup-details}
\begin{wrapfigure}{r}{0.3\textwidth}
    \centering
    \includegraphics[width=0.3\textwidth]{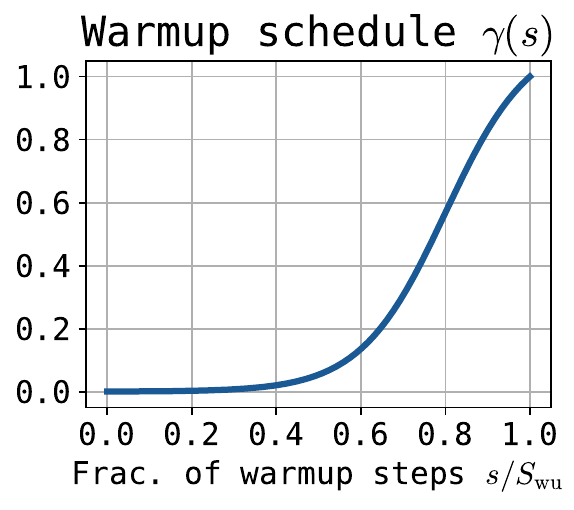}
    \caption{Diffusion-to-encoder warmup schedule. The coefficient $\gamma(s)$ controls the strength of the diffusion gradient passed to the encoder during the first $S_{\mathrm{wu}}$ training steps.}
    \label{fig:warmup_schedule}
\end{wrapfigure}
This appendix gives the implementation details for the diffusion-to-encoder warmup used in \cref{sec:diffusion-to-encoder-warmup}.
We warm up only the gradient of the diffusion objective with respect to the latent encoder output.
This is implemented with a stop-gradient interpolation:
\begin{equation}
\label{eq:warmup_latent}
\mathbf{z}_0^{\mathrm{diff}}(s)
=
\gamma(s)\,\mathbf{z}_0
+
\bigl(1-\gamma(s)\bigr)\,\operatorname{sg}(\mathbf{z}_0),
\end{equation}
where $\operatorname{sg}(\cdot)$ denotes the stop-gradient operator and $s$ is the current training step.
In the forward pass, $\mathbf{z}_0^{\mathrm{diff}}(s)=\mathbf{z}_0$, so the diffusion branch always receives the current latent representation.
In the backward pass, the gradient with respect to $\mathbf{z}_0$ is multiplied by $\gamma(s)$.
Thus, the diffusion target is unchanged; only the strength with which $\mathcal{L}_{\mathrm{diff}}$ updates the latent encoder is controlled.

We define $\gamma(s)$ with a normalized sigmoid schedule.
Let
\begin{equation}
\label{eq:warmup_sigmoid}
\tilde{\sigma}(s)
=
\sigma\!\left(
k \left(\frac{s}{S_{\mathrm{wu}}} - c\right)
\right),
\end{equation}
where $\sigma(\cdot)$ is the logistic sigmoid, $S_{\mathrm{wu}}$ is the warmup length, $c$ controls the midpoint of the rise, and $k$ controls the steepness.
We then set
\begin{equation}
\label{eq:warmup_gamma}
\gamma(s)
=
\begin{cases}
\gamma_{\min}
+
(1-\gamma_{\min})
\displaystyle
\frac{
\tilde{\sigma}(s)-\tilde{\sigma}(0)
}{
\tilde{\sigma}(S_{\mathrm{wu}})-\tilde{\sigma}(0)
},
& 0 \le s \le S_{\mathrm{wu}}, \\[1.2em]
1, & s > S_{\mathrm{wu}}.
\end{cases}
\end{equation}
In all experiments, we use $k=10$ and $c=0.8$, which keeps $\gamma(s)$ close to $\gamma_{\min}$ at the beginning of training and increases $\gamma(s)$ smoothly towards $1$ near the end of the warmup period. This prevents the early, poorly fitted diffusion prior from strongly affecting the encoder before the reconstruction space becomes stable.
We illustrate this schedule in \cref{fig:warmup_schedule}.



\section{Ablations of framework architecture}\label{app:architecture_ablations}

\subsection{Effect of GPT-2 layer index}
\label{app:gpt_layers}

In this section, we study how the choice of GPT-2 layer affects generation quality. 
We compare three methods that operate on contextual GPT-2 representations: TEncDM, a COSMOS-style smoothed latent space, and our jointly trained model. 
The layer index specifies which hidden states of the frozen GPT-2 encoder are used as input to the latent encoder $\mathcal{E}^{\theta}_\mathbf{z}$.
For our method, the same hidden states are also used as the reconstruction target in $\mathcal{L}_\mathbf{h}$.
All models are trained on OWT-128 with the same diffusion backbone and decoder.

The results are shown in \cref{tab:gpt_layer_sweep}. 
For TEncDM, generation quality changes only moderately across layers. 
The final GPT-2 layer performs worst in terms of generative perplexity, while earlier layers give slightly better results, but the overall difference remains limited.

For the COSMOS-style latent space, the choice of layer has a larger effect. 
The best result is obtained with the third-to-last layer, while the final layer gives substantially worse Mauve. 
This suggests that smoothing and robustness objectives are sensitive to the structure of the pretrained representation used as input.

Our model outperforms both baselines for every GPT-2 layer. 
It also does not show the same degradation on the final layer: using the last GPT-2 layer gives the lowest generative perplexity and a Mauve close to the best configuration. 
This indicates that the trainable latent encoder can reshape different GPT-2 hidden states into representations that are more suitable for diffusion.

\begin{table*}[t]
    \centering
    \small
    \caption{
    Effect of the GPT-2 hidden layer index on generation quality.
    All models are trained on OWT-128 with the same diffusion backbone and decoder.
    }
    \label{tab:gpt_layer_sweep}
    \setlength{\tabcolsep}{2.7pt}
    \begin{tabular}{llcccc}
    \toprule
    Method & Layer & Gen. PPL ($\downarrow$) & Mauve ($\uparrow$) & Div ($\uparrow$) & Entropy ($\uparrow$) \\
    \midrule
    \multirow{4}{*}{TEncDM}
    & $-1$ & $158.4_{\pm 1.3}$ & $25.5_{\pm 4.4}$ & $54.7_{\pm 0.4}$ & $4.26_{\pm 0.01}$ \\
    & $-2$ & $156.5_{\pm 2.2}$ & $33.3_{\pm 3.4}$ & $54.7_{\pm 0.4}$ & $4.27_{\pm 0.00}$ \\
    & \cellcolor{gray!20}$\boldsymbol{-3}$ & \cellcolor{gray!20}$\mathbf{149.7}_{\pm 1.8}$ & \cellcolor{gray!20}$36.6_{\pm 5.9}$ & \cellcolor{gray!20}$55.4_{\pm 0.3}$ & \cellcolor{gray!20}$\mathbf{4.28}_{\pm 0.00}$ \\
    & $-4$ & $154.4_{\pm 1.8}$ & $\mathbf{37.6}_{\pm 5.1}$ & $\mathbf{56.6}_{\pm 0.4}$ & $4.27_{\pm 0.00}$ \\

    \midrule
    \multirow{4}{*}{COSMOS-style}
    & $-1$ & $137.6_{\pm 0.7}$ & $9.6_{\pm 1.5}$ & $50.0_{\pm 0.2}$ & $4.11_{\pm 0.01}$ \\
    & $-2$ & $105.3_{\pm 2.1}$ & $48.1_{\pm 4.2}$ & $50.4_{\pm 0.3}$ & $4.17_{\pm 0.01}$ \\
    & \cellcolor{gray!20}$\boldsymbol{-3}$ & \cellcolor{gray!20}$\mathbf{98.5}_{\pm 1.8}$ & \cellcolor{gray!20}$\mathbf{53.4}_{\pm 4.8}$ & \cellcolor{gray!20}$51.0_{\pm 0.4}$ & \cellcolor{gray!20}$\mathbf{4.19}_{\pm 0.01}$ \\
    & $-4$ & $116.1_{\pm 1.6}$ & $37.7_{\pm 2.0}$ & $\mathbf{53.3}_{\pm 0.3}$ & $4.15_{\pm 0.01}$ \\
    \midrule
    \multirow{4}{*}{\textbf{LDLM (Ours)}}
    & \cellcolor{gray!20}$\boldsymbol{-1}$ & \cellcolor{gray!20}$\mathbf{54.5}_{\pm 0.8}$ & \cellcolor{gray!20}$88.1_{\pm 0.9}$ & \cellcolor{gray!20}$52.1_{\pm 0.5}$ & \cellcolor{gray!20}$4.24_{\pm 0.00}$ \\
    & $-2$ & $72.7_{\pm 0.4}$ & $87.2_{\pm 1.8}$ & $\mathbf{53.8}_{\pm 0.2}$ & $\mathbf{4.26}_{\pm 0.01}$ \\
    & $-3$ & $66.3_{\pm 0.8}$ & $\mathbf{89.1}_{\pm 3.1}$ & $52.9_{\pm 0.3}$ & $4.25_{\pm 0.01}$ \\
    & $-4$ & $63.8_{\pm 0.6}$ & $82.5_{\pm 2.6}$ & $51.7_{\pm 0.3}$ & $4.24_{\pm 0.00}$ \\
    \bottomrule
    \end{tabular}
\end{table*}

\subsection{Post-training a smaller token decoder}
\label{app:decoder_post_training}

In this section, we study whether the decoder can be reduced after the latent encoder and diffusion model have been trained. 
The motivation is that, during joint training, the decoder is used not only to predict tokens but also to reconstruct hidden states of the frozen semantic encoder. 
This hidden-state reconstruction is important for shaping the latent space, but after the encoder and diffusion are fixed, the final inference objective is only to map generated latents to tokens.

\begin{table*}[h]
    \centering
    \small
    \caption{
    Effect of the number of post-trained token decoder layers on generation quality.
    The encoder and diffusion model are fixed.
    Token decoders are trained on noisy latents with $\sigma_{\mathrm{dec}}=3.0$ using only cross-entropy.
    }
    \label{tab:decoder_post_training}
    \setlength{\tabcolsep}{2.7pt}
    \begin{tabular}{lcccc}
    \toprule
    Decoder configuration & Gen. PPL ($\downarrow$) & Mauve ($\uparrow$) & Div ($\uparrow$) & Entropy ($\uparrow$) \\
    \midrule
    Reference jointly trained decoder
    & $66.3_{\pm 0.8}$ & $\mathbf{89.1}_{\pm 3.1}$ & $\mathbf{52.9}_{\pm 0.3}$ & $4.25_{\pm 0.01}$ \\
    \midrule
    1 layer
    & $81.3_{\pm 1.2}$ & $85.9_{\pm 2.2}$ & $52.0_{\pm 0.3}$ & $\mathbf{4.26}_{\pm 0.00}$ \\
    2 layers
    & $68.7_{\pm 0.6}$ & $87.2_{\pm 2.9}$ & $52.0_{\pm 0.2}$ & $4.24_{\pm 0.00}$ \\
    3 layers
    & $67.0_{\pm 0.8}$ & $86.6_{\pm 2.5}$ & $51.4_{\pm 0.4}$ & $4.22_{\pm 0.00}$ \\
    6 layers
    & $\mathbf{65.1}_{\pm 0.5}$ & $86.2_{\pm 1.0}$ & $51.9_{\pm 0.1}$ & $4.23_{\pm 0.00}$ \\
    \bottomrule
    \end{tabular}
\end{table*}

We therefore train new token decoders on top of a fixed pretrained encoder and diffusion model to decrease the total number of parameters. 
Each decoder consists of a small number of transformer layers followed by a linear head. 
The decoder is trained on noisy latents using only the token-level cross-entropy loss, since hidden-state reconstruction is no longer needed in this post-training stage. 
We use the same decoder input noise as in the final model, with $\sigma_{\mathrm{dec}} = 3.0$.
All post-trained decoders are trained for 200{,}000 iterations with batch size 512.

In the standard jointly trained model, the decoder contains 9 transformer layers in total: 6 layers for hidden-state decoding and 3 layers for token decoding. 
In this experiment, we allocate all decoder capacity directly to token prediction and vary the number of transformer layers from 1 to 6.

The results are shown in \cref{tab:decoder_post_training}. 
A 1-layer decoder is clearly weaker than the reference jointly trained decoder. 
However, with 2 or more layers, the post-trained decoder reaches comparable generation quality. 
Increasing the number of layers further gives only a moderate improvement in generative perplexity.

These results indicate that a large decoder is mainly useful during joint training, where it helps train the latent space through hidden-state reconstruction. 
After the encoder and diffusion model are fixed, the decoder can be trained directly for token prediction and made substantially smaller without a large loss in generation quality.

\subsection{Pretrained encoder generalization}
\label{app:pretrained_encoder}

In this section, we test whether our approach generalizes to other pretrained semantic representations. 
We compare two frozen encoders: GPT-2 small and BERT-base~\citep{devlin2019bert}. 
For BERT-base, we use the same training recipe as for GPT-2: hidden-state reconstruction with MSE, diffusion-to-encoder warmup of 10k steps, adaptive timestep sampling, and $\sigma_{\mathrm{dec}} = 3.0$. We use the output of the third-to-last layer for BERT as input to the latent encoder.
We do not tune hyperparameters for BERT-base and keep the values that worked well for the GPT-2 model.

\begin{table*}[h]
    \centering
    \small
    \caption{
    Effect of the pretrained semantic encoder on generation quality.
    The BERT-base model uses the same training recipe as the GPT-2 small model, without additional hyperparameter tuning.
    }
    \label{tab:bert_results}
    \setlength{\tabcolsep}{2.7pt}
    \begin{tabular}{lcccc}
    \toprule
    Pretrained encoder & Gen. PPL ($\downarrow$) & Mauve ($\uparrow$) & Div ($\uparrow$) & Entropy ($\uparrow$) \\
    \midrule
    \rowcolor{gray!20} \textbf{GPT-2 small}
    & $\mathbf{66.3}_{\pm 0.8}$ & $89.1_{\pm 3.1}$ & $\mathbf{52.9}_{\pm 0.3}$ & $4.25_{\pm 0.01}$ \\
    BERT-base
    & $86.0_{\pm 0.3}$ & $\mathbf{93.6}_{\pm 6.3}$ & $47.8_{\pm 0.2}$ & $\mathbf{4.29}_{\pm 0.00}$ \\
    \bottomrule
    \end{tabular}
\end{table*}

The results are shown in \cref{tab:bert_results}. 
BERT-base still gives strong generation quality under the same training recipe. 
Overall, the model remains competitive without any BERT-specific hyperparameter tuning.

These results suggest that the proposed joint training approach is not tied to GPT-2 representations. 
It can also be applied to other pretrained semantic encoders, although the best hyperparameters may depend on the choice of encoder.

\subsection{Latent encoder architecture}
\label{app:latent_encoder_architecture}

In this section, we compare two architectures for the latent encoder and hidden-state decoder. 
Our default model uses a Perceiver Resampler-style encoder~\citep{flamingo}, which has also been applied in prior latent text diffusion models~\citep{cosmos,ld4lg}. 
This architecture maintains a set of trainable latent vectors and updates them through cross-attention to the pre-trained hidden states. 
Thus, the latent vectors directly aggregate information from the full input sequence, while the pre-trained hidden states remain fixed across encoder layers.

\begin{table*}[h]
    \centering
    \small
    \caption{
    Effect of the latent encoder architecture on generation quality.
    }
    \label{tab:latent_encoder_architecture}
    \setlength{\tabcolsep}{2.7pt}
    \begin{tabular}{lcccc}
    \toprule
    Latent encoder & Gen. PPL ($\downarrow$) & Mauve ($\uparrow$) & Div ($\uparrow$) & Entropy ($\uparrow$) \\
    \midrule
    \rowcolor{gray!20} \textbf{Perceiver Resampler}
    & $\mathbf{66.3}_{\pm 0.8}$ & $89.1_{\pm 3.1}$ & $\mathbf{52.9}_{\pm 0.3}$ & $\mathbf{4.25}_{\pm 0.01}$ \\
    Self-attention
    & $94.7_{\pm 1.4}$ & $\mathbf{91.9}_{\pm 0.4}$ & $49.6_{\pm 0.4}$ & $4.21_{\pm 0.00}$ \\
    \bottomrule
    \end{tabular}
\end{table*}

We compare this design with a bidirectional self-attention encoder.
In this variant, the pre-trained hidden states are processed by standard transformer blocks, and the resulting contextual states are used as latents for diffusion. 
Unlike the Perceiver Resampler, this architecture forms latents by repeatedly updating the pre-trained hidden states themselves through self-attention.

Both models are trained on OWT-128 with the same recipe: hidden-state reconstruction with MSE, diffusion-to-encoder warmup of 10k steps, adaptive timestep sampling, and $\sigma_{\mathrm{dec}} = 3.0$. 
We use the third-to-last GPT-2 layer as input to the latent encoder.

The results are shown in \cref{tab:latent_encoder_architecture}. 
The self-attention encoder gives reasonable generation quality, which indicates that the proposed training recipe is not restricted to the Perceiver Resampler architecture. 
However, it performs worse than the Perceiver Resampler in generative perplexity and diversity. 
This suggests that explicitly aggregating information into trainable latent vectors through cross-attention is a better fit for our latent diffusion setup than directly updating pre-trained hidden states with self-attention.


\clearpage
\section{Latent space smoothness}
\label{app:latent_smoothness}

Following prior work on latent text diffusion~\citep{planner,cosmos}, we use interpolation smoothness as a proxy for how well a latent space suits diffusion: in a smooth space, points between the latents of two real texts still decode into fluent text, so small denoising errors are less likely to break the generated text.

We compare the TEncDM, Frozen enc., COSMOS-style, and LDLM latent spaces from \cref{sec:space_comparison}. All of them are built on top of the same frozen GPT-2 token encoder and are evaluated on the same 100 random pairs of OWT-128 validation texts. For each pair, we encode both texts and linearly interpolate between their latents, $\mathbf{z}_\alpha=(1-\alpha)\mathbf{z}_1+\alpha\mathbf{z}_2$, at 101 equally spaced values of $\alpha\in[0,1]$. We decode every point with the decoder of the corresponding model using greedy token selection, without diffusion denoising, and compute Gen. PPL of the decoded texts with GPT-2 Large.

\begin{figure}[ht]
    \centering
    \includegraphics[width=0.8\textwidth]{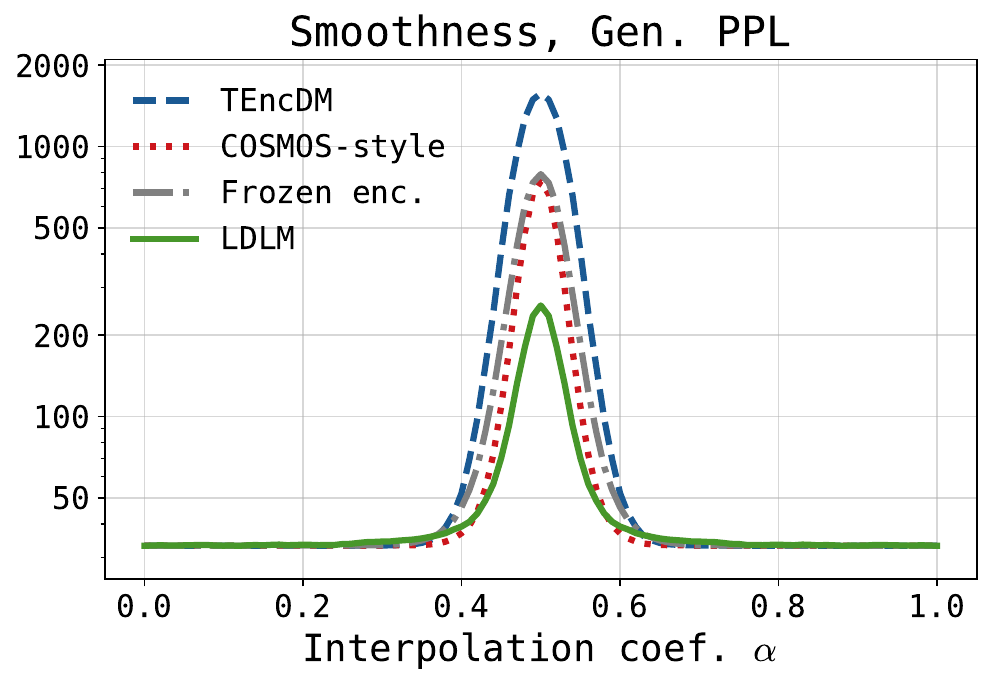}
    \caption{Latent space smoothness: Gen. PPL (log scale) of texts decoded from linear interpolations between the latents of two OWT-128 validation texts, averaged over 100 pairs. Lower is smoother. All models are evaluated on the same text pairs.}
    \label{fig:space_smoothness}
\end{figure}

\cref{fig:space_smoothness} shows the results. At the endpoints, all curves coincide, as every model reconstructs real texts accurately; the differences appear in the middle of the path, where the latents are farthest from real data. The TEncDM space degrades the most. The Frozen enc. and COSMOS-style spaces are noticeably smoother, with the explicit smoothing augmentations of COSMOS-style giving a small additional gain. The jointly trained LDLM space is the smoothest by a wide margin: its peak Gen. PPL is roughly $3\times$ lower than that of Frozen enc., which differs from LDLM only in that the diffusion gradient does not reach the latent encoder.

\section{Implementation details}
\label{app:impl_det}

\paragraph{Optimization.}
We train all models with AdamW using a learning rate $3\cdot 10^{-4}$, weight decay $0.01$, $\beta_1=0.9$, $\beta_2=0.98$, and $\epsilon=10^{-8}$.
The learning rate is linearly warmed up from $10^{-8}$ to $3\cdot 10^{-4}$ during the first 5,000 iterations and then kept constant.
We clip the gradient norm to $1.0$.
All ablation models are trained for 500,000 iterations with batch size 512.
For the main LM1B and OpenWebText experiments, we train for 1,000,000 iterations with batch size 512.
All LM1B baselines are trained for the same number of iterations and with the same batch size.

\paragraph{Architecture.}
Unless stated otherwise, we use GPT-2 small as the frozen token encoder and take hidden states from the third-to-last layer.
The latent dimensionality is $768$, matching the hidden size of GPT-2 small.
The number of latent vectors is equal to the sequence length: 128 for LM1B and OWT-128 ablations, and 1024 for the main OpenWebText experiments.
Both the latent encoder and latent decoder use 6-layer Perceiver Resamplers with hidden size $768$ and 12 attention heads.
The diffusion model is a 12-layer Diffusion Transformer with hidden size $768$ and 12 attention heads.
We use self-conditioning with probability $0.5$.

\paragraph{Diffusion and sampling.}
We train the diffusion model with $x_0$-prediction.
For the forward process, we use the tangent noise schedule with $d=3$ from COSMOS~\citep{cosmos}.
At sampling time, we use the Euler-Maruyama solver.

\paragraph{Training recipe.}
For LM1B and OpenWebText, we use a diffusion-to-encoder warmup of $S_{\mathrm{wu}}=50\mathrm{k}$ steps, decoder-input noise $\sigma_{\mathrm{dec}}=3$, hidden-state MSE reconstruction, and adaptive timestep sampling.
For ablations, we use the settings stated in \cref{sec:ablations} unless the ablated component is changed.

\paragraph{Data preprocessing.}
We follow the preprocessing protocol of MDLM~\citep{mdlm}.
Texts are tokenized with the GPT-2 tokenizer, concatenated into a single token stream, and split into fixed-length chunks.
We do not use padding.

\paragraph{Compute.}
The main LM1B run takes approximately 5.5 days on 4 NVIDIA A100 GPUs.
The main OpenWebText run takes approximately 5 days on 64 NVIDIA A100 GPUs.
For timing comparisons, we measure sampling time on OpenWebText with sequence length 1024, batch size 16, and 1024 denoising steps, using the same hardware setup for all methods.

\paragraph{Parameter counts.}
All parameter counts in this paragraph refer to trainable parameters and exclude the frozen GPT-2 token encoder.
The latent encoder and latent decoder contain about 50M parameters each, and the denoiser contains 132M parameters.
The token decoder size depends on its depth: the 3-layer Transformer decoder used in our main experiments contains 66M parameters including the vocabulary projection, while a linear-head-only decoder contains 38M parameters.
At inference time, neither the frozen token encoder nor the trainable latent encoder is used, since generation starts directly from Gaussian latent noise.
During iterative sampling, LDLM repeatedly evaluates only the denoiser, which has no token embedding matrix and no vocabulary-sized output head.
The latent decoder and the token decoder, including the expensive vocabulary projection, are applied only once after denoising is complete.
This substantially reduces the cost of sampling compared to token-level diffusion models that project to the vocabulary at every denoising step.
As shown in \cref{app:decoder_post_training}, after the latent encoder and denoiser are fixed, the token decoder can be trained with cross-entropy and reduced to a linear head.

\section{Detailed unconditional generation results}
\label{sec:full_comparison_tables}

This section reports the full results that support the
quality-diversity Pareto curves shown in \cref{sec:main_results}.
\Cref{tab:lm1b_results} contains the LM1B benchmark with sequence length~128
and \cref{tab:owt1024_results} contains the OpenWebText benchmark with
sequence length~1024.
For every method we report the four evaluation metrics introduced in
\cref{sec:ablations}: generative perplexity, token-level entropy,
$n$-gram diversity, and Mauve across sampling budgets ranging from
$32$ to $4096$ denoising steps for OpenWebText and from $32$ to $1024$ for LM1B,
so that low-step and high-step regimes can be compared at a glance.
The set of baselines matches the one used throughout the main text
(MDLM, MDLM with ReMDM remasking under the \emph{rescale} schedule, DUO with and without the $\Psi$-sampler, CANDI, and FLM),
and our model is reported in the last group as LDLM.

\begin{table}[t]
    \small
    \centering
    \setlength{\tabcolsep}{2.8pt}
    \caption{Unconditional generation results on \textbf{LM1B} (sequence length~128) across a range of sampling budgets. Mauve and Div are reported as percentages.}
    \label{tab:lm1b_results}
    \resizebox{\textwidth}{!}{%
    \begin{tabular}{l l c c c c c c}
    \toprule
    Method & Metric & \multicolumn{6}{c}{Steps (NFE)} \\
    \cmidrule(lr){3-8}
                    &                 & $32$ & $64$ & $128$ & $256$ & $512$ & $1024$ \\
    \midrule
    \multirow{4}{*}{Real texts} & Gen.\ PPL $\downarrow$ & \multicolumn{6}{c}{$40.2$} \\
     & Mauve $\uparrow$ & \multicolumn{6}{c}{$100.0$} \\
     & Div $\uparrow$ & \multicolumn{6}{c}{$62.3$} \\
     & Ent $\uparrow$ & \multicolumn{6}{c}{$4.37$} \\
    \midrule
    \multirow{4}{*}{MDLM~\citep{mdlm}}
    & Gen.\ PPL $\downarrow$
    & $196.9_{\pm2.1}$ & $161.2_{\pm1.2}$ & $139.2_{\pm1.3}$ & $123.8_{\pm.6}$ & $109.7_{\pm.8}$ & $97.5_{\pm.8}$ \\
    & Mauve $\uparrow$
    & $73.3_{\pm1.9}$ & $79.6_{\pm3.4}$ & $89.1_{\pm4.2}$ & $89.3_{\pm4.8}$ & $88.6_{\pm3.0}$ & $90.1_{\pm1.8}$ \\
    & Div $\uparrow$
    & $66.1_{\pm.1}$ & $64.5_{\pm.2}$ & $63.0_{\pm.2}$ & $61.4_{\pm.3}$ & $59.7_{\pm.2}$ & $57.8_{\pm.3}$ \\
    & Ent $\uparrow$
    & $4.38_{\pm.00}$ & $4.38_{\pm.00}$ & $4.37_{\pm.00}$ & $4.36_{\pm.00}$ & $4.35_{\pm.00}$ & $4.34_{\pm.00}$ \\
    \midrule
    \multirow{4}{*}{\makecell[l]{ReMDM (rescale)~\citep{wang2025remasking} \\ $p=0.9$}}
    & Gen.\ PPL $\downarrow$
    & $148.4_{\pm1.3}$ & $117.4_{\pm1.4}$ & $98.7_{\pm1.3}$ & $84.3_{\pm1.1}$ & $73.7_{\pm.8}$ & $64.8_{\pm1.0}$ \\
    & Mauve $\uparrow$
    & $84.2_{\pm4.6}$ & $88.5_{\pm1.8}$ & $91.9_{\pm1.6}$ & $90.4_{\pm2.4}$ & $91.6_{\pm1.6}$ & $91.5_{\pm1.3}$ \\
    & Div $\uparrow$
    & $63.5_{\pm.3}$ & $62.1_{\pm.4}$ & $60.9_{\pm.2}$ & $59.6_{\pm.3}$ & $58.3_{\pm.1}$ & $57.2_{\pm.2}$ \\
    & Ent $\uparrow$
    & $4.36_{\pm.00}$ & $4.36_{\pm.00}$ & $4.35_{\pm.00}$ & $4.35_{\pm.00}$ & $4.34_{\pm.00}$ & $4.34_{\pm.00}$ \\
    \midrule
    \multirow{4}{*}{Duo~\citep{duo}}
    & Gen.\ PPL $\downarrow$
    & $148.4_{\pm1.2}$ & $139.5_{\pm1.7}$ & $136.3_{\pm2.0}$ & $133.1_{\pm2.1}$ & $132.7_{\pm1.0}$ & $130.8_{\pm.8}$ \\
    & Mauve $\uparrow$
    & $84.5_{\pm2.8}$ & $88.8_{\pm2.1}$ & $90.6_{\pm1.5}$ & $90.4_{\pm2.0}$ & $91.7_{\pm1.5}$ & $88.9_{\pm1.3}$ \\
    & Div $\uparrow$
    & $62.7_{\pm.1}$ & $62.7_{\pm.3}$ & $62.9_{\pm.3}$ & $62.7_{\pm.3}$ & $63.0_{\pm.1}$ & $62.7_{\pm.1}$ \\
    & Ent $\uparrow$
    & $4.36_{\pm.00}$ & $4.37_{\pm.00}$ & $4.36_{\pm.00}$ & $4.37_{\pm.00}$ & $4.37_{\pm.00}$ & $4.36_{\pm.00}$ \\
    \midrule
    \multirow{4}{*}{\makecell[l]{Duo$\,$+$\,\Psi$-sampler~\citep{duo_ch2} \\ $p=0.9$ \\ $\eta=0.05$}}
    & Gen.\ PPL $\downarrow$
    & $75.0_{\pm.6}$ & $71.1_{\pm.4}$ & $68.4_{\pm.5}$ & $64.1_{\pm.9}$ & $60.3_{\pm.4}$ & $56.2_{\pm.3}$ \\
    & Mauve $\uparrow$
    & $89.6_{\pm1.5}$ & $92.0_{\pm1.2}$ & $92.7_{\pm.5}$ & $92.0_{\pm1.3}$ & $93.3_{\pm1.1}$ & $92.6_{\pm1.3}$ \\
    & Div $\uparrow$
    & $56.2_{\pm.2}$ & $56.5_{\pm.2}$ & $56.3_{\pm.1}$ & $56.1_{\pm.2}$ & $55.7_{\pm.2}$ & $55.2_{\pm.3}$ \\
    & Ent $\uparrow$
    & $4.32_{\pm.00}$ & $4.32_{\pm.00}$ & $4.33_{\pm.00}$ & $4.32_{\pm.00}$ & $4.33_{\pm.00}$ & $4.32_{\pm.00}$ \\
    \midrule
    \multirow{4}{*}{CANDI~\citep{candi}}
    & Gen.\ PPL $\downarrow$
    & $235.8_{\pm1.8}$ & $199.2_{\pm3.0}$ & $182.3_{\pm2.9}$ & $173.9_{\pm1.2}$ & $172.8_{\pm1.5}$ & $167.9_{\pm3.0}$ \\
    & Mauve $\uparrow$
    & $81.6_{\pm3.1}$ & $86.4_{\pm3.1}$ & $88.5_{\pm1.3}$ & $87.7_{\pm2.4}$ & $87.9_{\pm2.9}$ & $89.0_{\pm1.9}$ \\
    & Div $\uparrow$
    & $67.5_{\pm.2}$ & $66.6_{\pm.3}$ & $65.9_{\pm.2}$ & $65.7_{\pm.3}$ & $65.8_{\pm.1}$ & $65.6_{\pm.2}$ \\
    & Ent $\uparrow$
    & $4.41_{\pm.00}$ & $4.41_{\pm.00}$ & $4.41_{\pm.00}$ & $4.41_{\pm.00}$ & $4.40_{\pm.00}$ & $4.40_{\pm.00}$ \\
    \midrule
    \multirow{4}{*}{FLM~\citep{lee2026flow}}
    & Gen.\ PPL $\downarrow$
    & $226.9_{\pm2.1}$ & $188.8_{\pm.9}$ & $167.8_{\pm1.8}$ & $154.5_{\pm.8}$ & $147.1_{\pm.8}$ & $142.5_{\pm.7}$ \\
    & Mauve $\uparrow$
    & $41.8_{\pm5.4}$ & $52.7_{\pm6.9}$ & $61.1_{\pm5.1}$ & $65.3_{\pm9.0}$ & $62.8_{\pm3.9}$ & $64.2_{\pm2.4}$ \\
    & Div $\uparrow$
    & $61.0_{\pm.1}$ & $58.6_{\pm.1}$ & $56.8_{\pm.1}$ & $55.7_{\pm.1}$ & $54.9_{\pm.2}$ & $54.5_{\pm.2}$ \\
    & Ent $\uparrow$
    & $4.45_{\pm.00}$ & $4.42_{\pm.00}$ & $4.40_{\pm.00}$ & $4.38_{\pm.00}$ & $4.37_{\pm.00}$ & $4.36_{\pm.00}$ \\
    \midrule
    \multirow{4}{*}{\textbf{LDLM (Ours)}}
    & Gen.\ PPL $\downarrow$
    & $154.1_{\pm1.6}$ & $99.1_{\pm1.5}$ & $79.2_{\pm.8}$ & $68.3_{\pm.6}$ & $63.0_{\pm.5}$ & $57.4_{\pm.8}$ \\
    & Mauve $\uparrow$
    & $75.2_{\pm2.1}$ & $93.7_{\pm1.1}$ & $93.8_{\pm1.1}$ & $92.7_{\pm2.0}$ & $91.3_{\pm1.9}$ & $89.2_{\pm1.2}$ \\
    & Div $\uparrow$
    & $67.3_{\pm.2}$ & $62.0_{\pm.2}$ & $58.8_{\pm.3}$ & $56.7_{\pm.2}$ & $55.4_{\pm.3}$ & $53.8_{\pm.2}$ \\
    & Ent $\uparrow$
    & $4.42_{\pm.00}$ & $4.39_{\pm.00}$ & $4.38_{\pm.00}$ & $4.37_{\pm.00}$ & $4.37_{\pm.00}$ & $4.36_{\pm.00}$ \\
    \bottomrule
    \end{tabular}%
    }
    \end{table}

\begin{table}[t]
    \small
    \centering
    \setlength{\tabcolsep}{2.8pt}
    \caption{Unconditional generation results on \textbf{OpenWebText} (sequence length~1024) across a range of sampling budgets. Mauve and Div are reported as percentages.}
    \label{tab:owt1024_results}
    \resizebox{\textwidth}{!}{%
    \begin{tabular}{llcccccccc}
    \toprule
    Method & Metric & \multicolumn{8}{c}{Steps (NFE)} \\
    \cmidrule(lr){3-10}
                    &                 & $32$ & $64$ & $128$ & $256$ & $512$ & $1024$ & $2048$ & $4096$ \\
    \midrule
    \multirow{4}{*}{Real texts} & Gen.\ PPL ($\downarrow$) & \multicolumn{8}{c}{$14.6$} \\
     & Mauve ($\uparrow$) & \multicolumn{8}{c}{$100.0$} \\
     & Div ($\uparrow$) & \multicolumn{8}{c}{$33.1$} \\
     & Ent ($\uparrow$) & \multicolumn{8}{c}{$5.436$} \\
     \midrule
    \multirow{4}{*}{MDLM~\citep{mdlm}} & Gen.\ PPL $\downarrow$ &
    $197_{\pm1.2}$ &
    $142_{\pm1.5}$ &
    $120_{\pm1.3}$ &
    $110_{\pm.7}$ &
    $107_{\pm1}$ &
    $104_{\pm1.1}$ &
    $104_{\pm.7}$ &
    $102_{\pm1}$ \\
    
    & Mauve $\uparrow$ &
    $0.7_{\pm.0}$ &
    $1_{\pm.14}$ &
    $1.5_{\pm.1}$ &
    $2.4_{\pm.6}$ &
    $2.7_{\pm.7}$ &
    $3.2_{\pm1.2}$ &
    $2.7_{\pm.4}$ &
    $3_{\pm.6}$ \\
    
    & Div $\uparrow$ &
    $45.4_{\pm.1}$ &
    $42.7_{\pm.2}$ &
    $41.4_{\pm.3}$ &
    $40.9_{\pm.1}$ &
    $41.0_{\pm.2}$ &
    $40.9_{\pm.2}$ &
    $41.0_{\pm.1}$ &
    $40.7_{\pm.2}$ \\
    
    & Ent $\uparrow$ &
    $5.75_{\pm.00}$ &
    $5.70_{\pm.00}$ &
    $5.67_{\pm.00}$ &
    $5.65_{\pm.00}$ &
    $5.64_{\pm.00}$ &
    $5.63_{\pm.00}$ &
    $5.64_{\pm.00}$ &
    $5.63_{\pm.00}$ \\
    \midrule
    \multirow{4}{*}{\makecell[l]{MDLM~\citep{mdlm} \\ $p=0.9$}}
    & Gen.\ PPL $\downarrow$
    & $69.21_{\pm.3}$
    & $51.15_{\pm.2}$
    & $43.69_{\pm.3}$
    & $39.50_{\pm.4}$
    & $38.05_{\pm.3}$
    & $37.52_{\pm.2}$
    & $37.16_{\pm.3}$
    & $36.4_{\pm.2}$ \\
    
    & Mauve $\uparrow$
    & $1.9_{\pm.3}$
    & $6.1_{\pm2.0}$
    & $11.7_{\pm4.2}$
    & $14.2_{\pm4.8}$
    & $20.1_{\pm5.5}$
    & $25.8_{\pm5.1}$
    & $22.6_{\pm2.1}$
    & $21.8_{\pm5.4}$ \\
    
    & Div $\uparrow$
    & $29.4_{\pm.1}$
    & $27_{\pm.1}$
    & $25.9_{\pm.2}$
    & $25_{\pm.3}$
    & $24.9_{\pm.1}$
    & $24.8_{\pm.1}$
    & $24.8_{\pm.2}$
    & $24.4_{\pm.2}$ \\
    
    & Ent $\uparrow$
    & $5.48_{\pm.00}$
    & $5.43_{\pm.00}$
    & $5.39_{\pm.00}$
    & $5.36_{\pm.00}$
    & $5.34_{\pm.01}$
    & $5.33_{\pm.01}$
    & $5.33_{\pm.00}$
    & $5.32_{\pm.00}$ \\
    \midrule
    \multirow{4}{*}{\makecell[l]{ReMDM (rescale)~\citep{wang2025remasking} \\ $p=0.9$}}
    & Gen.\ PPL $\downarrow$
    & $69.1_{\pm.6}$
    & $50.4_{\pm.5}$
    & $41.9_{\pm.3}$
    & $37.2_{\pm.6}$
    & $33.7_{\pm.3}$
    & $30.1_{\pm.3}$
    & $25.8_{\pm.2}$
    & $21.3_{\pm.1}$ \\
    
    & Mauve $\uparrow$
    & $1.5_{\pm.1}$
    & $6.5_{\pm2.5}$
    & $15.4_{\pm4.3}$
    & $21.2_{\pm7.5}$
    & $29.3_{\pm8.5}$
    & $36.7_{\pm13.4}$
    & $47.8_{\pm5.9}$
    & $46.4_{\pm5.9}$ \\
    
    & Div $\uparrow$
    & $29.2_{\pm.2}$
    & $26.6_{\pm.2}$
    & $25.5_{\pm.3}$
    & $24.8_{\pm.4}$
    & $23.8_{\pm.3}$
    & $23.3_{\pm.2}$
    & $21.6_{\pm.4}$
    & $19.3_{\pm.2}$ \\
    
    & Ent $\uparrow$
    & $5.49_{\pm.01}$
    & $5.43_{\pm.00}$
    & $5.38_{\pm.00}$
    & $5.35_{\pm.00}$
    & $5.33_{\pm.00}$
    & $5.29_{\pm.00}$
    & $5.24_{\pm.01}$
    & $5.17_{\pm.01}$ \\
    \midrule
    \multirow{4}{*}{Duo~\citep{duo}}
    & Gen.\ PPL $\downarrow$
    & $95.5_{\pm.8}$
    & $85.6_{\pm.3}$
    & $80.8_{\pm.6}$
    & $78.0_{\pm.4}$
    & $77.4_{\pm1.1}$
    & $76.5_{\pm.3}$
    & $76.0_{\pm1.3}$
    & $75.8_{\pm.8}$ \\
    
    & Mauve $\uparrow$
    & $1.0_{\pm.4}$
    & $4.0_{\pm1.3}$
    & $6.0_{\pm2.0}$
    & $5.0_{\pm1.0}$
    & $7.0_{\pm1.0}$
    & $7.0_{\pm1.0}$
    & $7.0_{\pm3.0}$
    & $5.8_{\pm2.0}$ \\
    
    & Div $\uparrow$
    & $35.9_{\pm.2}$
    & $35.9_{\pm.1}$
    & $36.2_{\pm.3}$
    & $36.2_{\pm.2}$
    & $36.4_{\pm.2}$
    & $36.6_{\pm.1}$
    & $36.4_{\pm.3}$
    & $36.4_{\pm.2}$ \\
    
    & Ent $\uparrow$
    & $5.57_{\pm.00}$
    & $5.57_{\pm.00}$
    & $5.56_{\pm.00}$
    & $5.55_{\pm.00}$
    & $5.55_{\pm.01}$
    & $5.54_{\pm.01}$
    & $5.54_{\pm.01}$
    & $5.53_{\pm.01}$ \\
    \midrule
    \multirow{4}{*}{\makecell[l]{Duo~\citep{duo_ch2} \\  $p=0.9$}}
    & Gen.\ PPL $\downarrow$
    & $44.22_{\pm.3}$
    & $39.71_{\pm.3}$
    & $37.97_{\pm.2}$
    & $36.54_{\pm.3}$
    & $36.16_{\pm.3}$
    & $35.64_{\pm.1}$
    & $35.58_{\pm.3}$
    & $35.37_{\pm.2}$ \\
    
    & Mauve $\uparrow$
    & $13.3_{\pm4.1}$
    & $24.2_{\pm8}$
    & $30.9_{\pm8.5}$
    & $36.5_{\pm5.2}$
    & $44.4_{\pm3.7}$
    & $34.5_{\pm8.4}$
    & $40.1_{\pm9.6}$
    & $45.9_{\pm8.6}$ \\
    
    & Div $\uparrow$
    & $26.1_{\pm.2}$
    & $26_{\pm.2}$
    & $26.5_{\pm.2}$
    & $26.6_{\pm.1}$
    & $26.7_{\pm.3}$
    & $26.8_{\pm.2}$
    & $26.7_{\pm.2}$
    & $26.9_{\pm.2}$ \\
    
    & Ent $\uparrow$
    & $5.416_{\pm.00}$
    & $5.40_{\pm.00}$
    & $5.40_{\pm.00}$
    & $5.38_{\pm.01}$
    & $5.37_{\pm.01}$
    & $5.37_{\pm.01}$
    & $5.37_{\pm.01}$
    & $5.36_{\pm.00}$ \\
    \midrule
    \multirow{4}{*}{\makecell[l]{Duo$\,$+$\,\Psi$-sampler~\citep{duo_ch2} \\  $p=0.9$ \\ $\eta=0.05$}}
    & Gen.\ PPL $\downarrow$
    & $43.70_{\pm.2}$
    & $38.68_{\pm.4}$
    & $36.00_{\pm.2}$
    & $33.32_{\pm.2}$
    & $30.59_{\pm.3}$
    & $27.08_{\pm.3}$
    & $22.55_{\pm.1}$
    & $18.30_{\pm.1}$ \\
    
    & Mauve $\uparrow$
    & $12_{\pm1.4}$
    & $22.8_{\pm9}$
    & $39.2_{\pm10}$
    & $41.1_{\pm7}$
    & $44.8_{\pm6.8}$
    & $61.1_{\pm7.7}$
    & $68.2_{\pm4.7}$
    & $76.5_{\pm4.7}$ \\
    
    & Div $\uparrow$
    & $25.9_{\pm.1}$
    & $25.8_{\pm.2}$
    & $26.1_{\pm.2}$
    & $25.7_{\pm.2}$
    & $25.1_{\pm.3}$
    & $24.2_{\pm.3}$
    & $22.1_{\pm.2}$
    & $19.4_{\pm.2}$ \\
    
    & Ent $\uparrow$
    & $5.41_{\pm.00}$
    & $5.40_{\pm.00}$
    & $5.39_{\pm.00}$
    & $5.37_{\pm.00}$
    & $5.35_{\pm.00}$
    & $5.31_{\pm.01}$
    & $5.26_{\pm.01}$
    & $5.20_{\pm.00}$ \\
    \midrule
    \multirow{4}{*}{CANDI~\citep{candi}}
    & Gen.\ PPL $\downarrow$
    & $51.1_{\pm.2}$
    & $42_{\pm.3}$
    & $36.8_{\pm.3}$
    & $42.3_{\pm.2}$
    & $33_{\pm.4}$
    & $39.2_{\pm.2}$
    & $39_{\pm.3}$
    & $38.8_{\pm.6}$ \\
    
    & Mauve $\uparrow$
    & $4_{\pm2}$
    & $8_{\pm2}$
    & $9.5_{\pm1}$
    & $11_{\pm3}$
    & $13.8_{\pm2}$
    & $13.8_{\pm1}$
    & $13.5_{\pm2}$
    & $13.6_{\pm2}$ \\
    
    & Div $\uparrow$
    & $23.6_{\pm0}$
    & $22.8_{\pm.3}$
    & $21.6_{\pm.2}$
    & $25.1_{\pm.1}$
    & $20.8_{\pm.2}$
    & $23.9_{\pm0}$
    & $24.2_{\pm.2}$
    & $24.2_{\pm.4}$ \\
    
    & Ent $\uparrow$
    & $5.27_{\pm.00}$
    & $5.23_{\pm.01}$
    & $5.17_{\pm.00}$
    & $5.23_{\pm.01}$
    & $5.10_{\pm.01}$
    & $5.17_{\pm.01}$
    & $5.17_{\pm.01}$
    & $5.16_{\pm.01}$ \\
    \midrule
    \multirow{4}{*}{FLM~\citep{lee2026flow}}
    & Gen.\ PPL $\downarrow$
    & $243.5_{\pm1.4}$
    & $149.0_{\pm1}$
    & $104.5_{\pm1.2}$
    & $82.9_{\pm.7}$
    & $70.7_{\pm.5}$
    & $63.0_{\pm.5}$
    & $58.2_{\pm.5}$
    & $55.6_{\pm.6}$ \\
    
    & Mauve $\uparrow$
    & $0.6_{\pm.0}$
    & $0.8_{\pm.1}$
    & $0.9_{\pm.1}$
    & $0.8_{\pm.1}$
    & $0.9_{\pm.1}$
    & $0.9_{\pm.1}$
    & $0.8_{\pm.1}$
    & $0.8_{\pm.1}$ \\
    
    & Div $\uparrow$
    & $43.5_{\pm.3}$
    & $33.9_{\pm.4}$
    & $27.5_{\pm.3}$
    & $23.7_{\pm.3}$
    & $21.2_{\pm.2}$
    & $19.4_{\pm.2}$
    & $18.3_{\pm.2}$
    & $17.6_{\pm.2}$ \\
    
    & Ent $\uparrow$
    & $5.74_{\pm.00}$
    & $5.69_{\pm.00}$
    & $5.59_{\pm.00}$
    & $5.50_{\pm.00}$
    & $5.41_{\pm.00}$
    & $5.35_{\pm.00}$
    & $5.30_{\pm.01}$
    & $5.27_{\pm.01}$ \\
    \midrule
    \multirow{4}{*}{\textbf{LDLM (Ours)}}
    & Gen.\ PPL $\downarrow$
    & $132.2_{\pm1.3}$
    & $53.8_{\pm.4}$
    & $41.6_{\pm.4}$
    & $34.3_{\pm.3}$
    & $30.2_{\pm.1}$
    & $27.8_{\pm.2}$
    & $26.1_{\pm.2}$
    & $25.0_{\pm.2}$ \\
    
    & Mauve $\uparrow$
    & $85.7_{\pm2.3}$
    & $89.3_{\pm.6}$
    & $91.1_{\pm2.0}$
    & $85.4_{\pm1.7}$
    & $84.7_{\pm1.8}$
    & $84.4_{\pm2.5}$
    & $85.7_{\pm3.1}$
    & $84.3_{\pm2.4}$ \\
    
    & Div $\uparrow$
    & $42.0_{\pm.2}$
    & $33.9_{\pm.3}$
    & $31.0_{\pm.2}$
    & $28.6_{\pm.2}$
    & $26.7_{\pm.2}$
    & $25.1_{\pm.3}$
    & $24.1_{\pm.2}$
    & $23.3_{\pm.2}$ \\
    
    & Ent $\uparrow$
    & $5.73_{\pm.00}$
    & $5.60_{\pm.00}$
    & $5.53_{\pm.01}$
    & $5.46_{\pm.01}$
    & $5.41_{\pm.01}$
    & $5.37_{\pm.01}$
    & $5.34_{\pm.01}$
    & $5.31_{\pm.01}$ \\
    \bottomrule
    \end{tabular}%
    }
    \end{table}

\section{About metrics}\label{app:metrics}

We report both entropy and diversity because they measure different types of variation in generated text. Entropy is computed for each generated sequence from its empirical token frequency distribution and is then averaged over samples. It therefore captures intra-sequence variability and penalizes repetitive outputs within a single text.

Diversity is computed over the full set of generated sequences:
\[
\mathrm{div}(y) =
\prod_{n=2}^{4}
\frac{\#\text{unique }n\text{-grams in } y}
{\#n\text{-grams in } y},
\]
where $y$ denotes the collection of generated texts. This metric captures corpus-level variation, i.e. how many distinct $n$-grams appear across different samples.

These metrics are not interchangeable. A model may generate high-entropy sequences while repeatedly sampling from the same local token distribution, resulting in limited corpus-level diversity. Conversely, a model may obtain high diversity by producing different repetitive patterns across samples, while each individual sequence remains low-entropy. Reporting both metrics helps distinguish these two failure modes and gives a more reliable view of the diversity of generated text.

\section{Existing assets and licenses}
\label{app:assets}

We use only publicly available datasets, pretrained models, and baseline implementations/checkpoints.
\cref{tab:assets} summarizes the external assets used in this work.

\begin{table}[H]
\small
\centering
\caption{External assets used in this work.}
\label{tab:assets}
\begin{tabular}{p{0.25\linewidth}p{0.25\linewidth}p{0.5\linewidth}}
\toprule
Asset & Use in this work & License / terms \\
\midrule
GPT-2~\citep{gpt2}
& Frozen token encoder and Gen. PPL evaluator
& Modified MIT License. \\

LM1B~\citep{chelba2013one}
& Generation benchmark
& Apache License 2.0. \\

OpenWebText~\citep{Gokaslan2019OpenWeb}
& Generation benchmark
& Dataset packaging released under CC0; original web text is not owned by the dataset creators. \\

MDLM~\citep{mdlm} / ReMDM~\citep{wang2025remasking} / Duo~\citep{duo} / FLM~\citep{lee2026flow}
& Baseline implementations and/or checkpoints
& Publicly released by the corresponding authors; licenses follow the respective repositories/checkpoint pages. \\
CANDI checkpoint~\citep{candi}
& Baseline evaluation on OpenWebText
& Non-public checkpoint provided by the authors for research evaluation. \\
\bottomrule
\end{tabular}
\end{table}
\section{Limitations}\label{app:limitations}

LDLM improves the quality-efficiency trade-off of diffusion-based text generation, but the extreme few-step regime remains challenging.
Our current sampler does not use distillation or specialized solver optimization, leaving room for further improvements at very small numbers of denoising steps.
We also use a simple training recipe and do not exhaustively tune all schedules, noise levels, or encoder/decoder architectures.
Future work could improve few-step sampling and further optimize the encoder and decoder without changing the core joint encoder-diffusion learning framework.

\section{Societal impact}\label{app:societal_impact}
LDLM could improve the efficiency and accessibility of non-autoregressive text generation. As with other text generation models, improved generation quality and lower sampling cost could also make it easier to produce misleading, low-quality, or spam-like text at scale. Responsible release should therefore consider evaluation of misuse risks and safeguards appropriate to the released artifacts.
\section{Generation samples}
We present unconditional generation examples on LM1B and OWT with corresponding NFEs of 128 and 1024. Each sample is annotated with its Gen. PPL
and token entropy.

\begin{figure}[p]
	\centering
	\begin{samplebox}{\normalfont\textbf{\textsc{LDLM} (Ours), Sampling Steps: 32} \hfill \normalfont\scriptsize \textcolor{darkgray}{Gen.PPL: \textbf{67.2} \,|\, Entropy: \textbf{4.40}}}
		\footnotesize\linespread{0.85}\selectfont
    \texttt{<|endoftext|>} with international aid -- and replacing 100 million to 300 million tons of spoiled food -- is an issue of concern.\texttt{<|endoftext|>}Maryland traffic was up more than 1,000 miles in July and its results the lowest since October 2005, when the railroad said its passengers were traveling just over 100 mph.\texttt{<|endoftext|>}At its current top tier of networks across 20 countries, NBC has 42 contract agreements covering 24 countries, including major broadcast networks like Fox and NBC, as well as the sports-rich "3" five, ESPN, and The Stanley Cup.\texttt{<|endoftext|>}She went missing in a parking lot at Virginia Tech without parents or staff since the disappearance of\texttt{<|endoftext|>}

	\end{samplebox}

	\begin{samplebox}{\normalfont\textbf{\textsc{LDLM} (Ours), Sampling Steps: 64} \hfill \normalfont\scriptsize \textcolor{darkgray}{Gen.PPL: \textbf{61.5} \,|\, Entropy: \textbf{4.51}}}
		\footnotesize\linespread{0.85}\selectfont
    \texttt{<|endoftext|>} passage of the No Child Left Behind Act, which provides equal grade representation in rural public schools, blocked a veto by 60 or more by the Senate at Wednesday.\texttt{<|endoftext|>}Arshes had no reason to worry about being denied a penalty kick after an incident on the left knee of Roma striker Carlos Tevez in last Wednesday's Champions League match against Shakhtar Donetsk.\texttt{<|endoftext|>}So how do we fix the almost-existent Twitter?\texttt{<|endoftext|>}The British Air Force (RAA) will cut flight capacity of 600 aircraft in the next three years by 25  .\texttt{<|endoftext|>}Since 2006, Salinao has hired a team of independent scientists to establish an\texttt{<|endoftext|>}

	\end{samplebox}

	\begin{samplebox}{\normalfont\textbf{\textsc{LDLM} (Ours), Sampling Steps: 128} \hfill \normalfont\scriptsize \textcolor{darkgray}{Gen.PPL: \textbf{38.1} \,|\, Entropy: \textbf{4.41}}}
		\footnotesize\linespread{0.85}\selectfont
    \texttt{<|endoftext|>}passage to the No Left Behind Act, which provides language instruction to financially disadvantaged charter schools, delayed passage by the House and Senate by five votes on Thursday.\texttt{<|endoftext|>}Mickelson was nothing but a shock focused on clinching her victory on a first-round clash with French Open champion David Nalbandian, when the second round of the Corona Classic was set to begin on Thursday.\texttt{<|endoftext|>}NEW YORK (Reuters) - Warner Brothers Co (DIS.N) will cut the payroll of 100 directors in the next three months to save money, a research group said on Monday as the entertainment giant's music division struggles to retain critical\texttt{<|endoftext|>}

	\end{samplebox}

	\begin{samplebox}{\normalfont\textbf{\textsc{LDLM} (Ours), Sampling Steps: 256} \hfill \normalfont\scriptsize \textcolor{darkgray}{Gen.PPL: \textbf{29.8} \,|\, Entropy: \textbf{4.34}}}
		\footnotesize\linespread{0.85}\selectfont
    \texttt{<|endoftext|>} they would be asked to vote here -- not Florida -- and they expect to have their delegates up in this state within or without the country stopping voting.\texttt{<|endoftext|>}Yet the American people know they don't like him.\texttt{<|endoftext|>}The new study shows that low-income teens may have a twice as high risk of that type of mental health problem as those who are developing multiple migraines; teenage adolescents are twice as likely to have an increased risk of developing post-traumatic stress disorder later in their lives, according to the study, published in 2009 in the Annals of the National Academy of Sciences.\texttt{<|endoftext|>}"I would say, 'This\texttt{<|endoftext|>}
	\end{samplebox}

	\begin{samplebox}{\normalfont\textbf{\textsc{LDLM} (Ours), Sampling Steps: 512} \hfill \normalfont\scriptsize \textcolor{darkgray}{Gen.PPL: \textbf{28.1} \,|\, Entropy: \textbf{4.32}}}
		\footnotesize\linespread{0.85}\selectfont
    \texttt{<|endoftext|>} expansion.\texttt{<|endoftext|>}Star Land Systems and Sun Systems continue to provide built-in services to customers and businesses in St. Louis County.\texttt{<|endoftext|>}Four people have been charged in connection with a fire at a nightclub in Melbourne which left six people dead and nearly 200 others missing.\texttt{<|endoftext|>}Germany's CAC dropped 2.2 per cent and France's FTSE, Europe's biggest stock index, lost 1.9 percent.\texttt{<|endoftext|>}Brendy DeMarcusne scored 14 points and A.J. Price added 14 points to lead UCLA (3-0), which won for only the third time in college history.\texttt{<|endoftext|>}Sotomayor met\texttt{<|endoftext|>}

	\end{samplebox}

	\begin{samplebox}{\normalfont\textbf{\textsc{LDLM} (Ours), Sampling Steps: 1024} \hfill \normalfont\scriptsize \textcolor{darkgray}{Gen.PPL: \textbf{24.8} \,|\, Entropy: \textbf{4.42}}}
		\footnotesize\linespread{0.85}\selectfont
    \texttt{<|endoftext|>}ROME (Reuters) - The U.N. Security Council will impose financial curbs after North Korea agreed on Thursday to suspend its nuclear program in return for a 25,000 euro reward meant to persuade it to halt uranium enrichment.\texttt{<|endoftext|>}Robinson's office: disappointing performance.\texttt{<|endoftext|>}According to latest data from the Organisation for Economic Cooperation and Development, Germany's GDP was set to shrank 0.2 \% in the third quarter after shrinking by 1.2 \% the previous quarter.\texttt{<|endoftext|>}Margise McKewan (Professor Lawton) said there were many reasons this would not have happened except for the contrast between the languages and\texttt{<|endoftext|>}

	\end{samplebox}
	\caption{Samples from \textsc{LDLM} trained on LM1B for varying number of steps.}
	\label{fig:qual_sample_fix_noise_ours_lm1b}
\end{figure}

\begin{figure}[p]
  \centering
  \begin{samplebox}{\normalfont\textbf{\textsc{LDLM} (Ours), Sampling Steps: 128} \hfill \normalfont\scriptsize \textcolor{darkgray}{Gen.PPL: \textbf{47.8} \,|\, Entropy: \textbf{5.52}}}
  \scriptsize\linespread{0.85}\selectfont
          \texttt{<|endoftext|>} of possession of an ounce or less of medical marijuana, the jury said.

    The couple were members of the family received different letters of support, the jury said in large part because of their illegal possession of small amounts of the drug.

    Under state law, Parker and McCormman received a minimum of five years in prison and a minimum of 10 years in prison for possession of up to 2,000 grams (8.5 ounces) of marijuana.

    Surlill's trial is scheduled for April at the Massachusetts 11th Court.\texttt{<|endoftext|>}Buy Photo The Gran's Chicken at 534 N. Harvan Ave. Thursday, February 17, 2016. (Photo: Romeo Poobman / Miami City Archives)

    John Herron (4th left), J.V. Mitchell (second left), Gillian Herst (3rd left) and Giuseppe Thonini (center) serve long-range restaurant The Gran's Chicken with Italian-Italian Chicken at 534 N. Harrington Ave. Thursday, February 17, 2016. (Photo: Rocky Poobman / Miami City Register)Buy Photo Image 1 of / 8 Caption Close The burglars: Watch carefully.

    The Italian-style theft that recently hit the front door of a City Hall restaurant returned to a humzied level on Thursday morning, with the glass-sted lights marking a dim doorway.

    STAND: GUIDE when waking up to explore the menu. Sto it with your own discretion -- especially if you're not planning one-day meals.

    "I've tried to get all things together," said 38-year-old MacGregson, who hails from East Miami earned her some kacolades for the fast-food restaurant that morphed into 40 rooms full of splithairs and tables. "They have enough cook staff, like, we all have."

    Italian-style chicken at 534 N. Harvan Ave. Thursday, February 17, 2016. (Photo: Vince Poobman / Miami City Archives)

    The restaurant offers nothing more than rich tomato-flavored pizzas, but regularly offers a menu besen with achevre-steak flavor. Customers can choose from anywhere from the spaghetti salad to modern cheesaccuros, while a generous selection of salads, kebabs and prosicoupes is nestled in a bar-like setting that lies like "a farm when it comes to fruits and vegetables."

    "I wouldn't really do that, because I owned the restaurant I stole," she said. "It's the money I make."

    But in a fast-food bar in downtown, where the \$20 million theft lies, she thinks it's an indversence.

    "I'm a thief here," her husband quipped. "Usually speaking they get requests, and maybe for a certain security reason we're trying to steal. If really it came like that I'd be with a credit card. That's what's what scares me. This is the kind of person that you'd call a thief."

    Matty Matthews, 30, a 32-year-old chef assistant at 401 N. South Shore Ave. at 401 N. 8th St., thinks the situation explains the restaurant is an indiversity.

    NEWSPORT Get the FL Breaking News newsletter delivered to your inbox We're sorry, something something went wrong Please try again soon, or contact Customer Service at 1-876-456. Delivery: Delivery: Fri Invalid email address Thank you! You're almost almost signed up for Rochester Breaking News Keep an eye out for an email to confirm your newsletter registration. More newsletters

    "I believe if it's doing something good then people will come that way," she said. But others "want to see it this way."

    On a Thursday morning inside a new brick-and-mortestoleer restaurant at 541 West East Ave. at N. Beach, Matthews greeted another chef offering fancgent 20th-century dishes.

    "I was always thinking, 'This is going to get better,'" said Steve McCormman, a 42-year-old former Hyde City resident who describes what her place is "cute rooms with a variety of ingredients" and is always hired to cook servers.

    At The Curtain, which is about a block's short walk from downtown, The Two Curtain opens; 10 a.m. to 11 p.m. -- and does free business from Mondays through Fridays.

    The Curtain opens 24-hours, 10 a.m. to 10 p.m. on Saturdays, Fridays, Fridays and Sundays.

    Mathews still credits the burglars of the past for establishing a reputation here. He says chefs who haven't let go of the multicolored cuisine that defines New York City now live near a Italian comforthouse: a perfectly-sealed space with marble access tables and Italian nachos.\texttt{<|endoftext|>}
  \end{samplebox}
  
  \begin{samplebox}{\normalfont\textbf{\textsc{LDLM} (Ours), Sampling Steps: 256} \hfill \normalfont\scriptsize \textcolor{darkgray}{Gen.PPL: \textbf{37.7} \,|\, Entropy: \textbf{5.40}}}
  \scriptsize\linespread{0.85}\selectfont
          \texttt{<|endoftext|>}'s an additional 20 miles north of Houston as a team of meteorologists concluded that rain is now expected to flood Amarillo for much of August. And the area of Texas' flood-ravtered region could rise to an all-time 100-degree high.

    Persant weather has been draging residents of Texas for much of the month, making it smart for any kind of rain. A cascade of snowstorms caused large numbers of homes to without power, draining one-third of many electric utilities in need of power. The storms have seen buildings and structures turn into small apartment rocks.

    PEARY TA, HOSTATOR:

    Damant rain has been flooding residents of Texas for much of the month, making it smart for any kind of rainfall.

    We're joined here in Camp Station, New Mexico, and our guest is Thom Greenin, medical professor at Texas State University. He's one of the authors of Texas Southern: A Look at Your State.

    THERAS GREENININ, BY BYLINE:

    A devastating burst of rain on the streets of Texas is raking down some buildings and structures and stripping into apartment rocks. They're terrible things. They're seriously changing. And conditions in Texas are getting so much worse. Texas is sliding into the most dangerous record-setting storm ever for anyone.

    MAVID EDRUNONUGH: I'm saying to a citizen 'My God, this is a huge storm and I'm not thinking about any storm. My God, it's a bad year.'

    1950 was a record-breaking pouring year, with more than 100,000 feet being covered in storms, 60 inches snow and nearly a ton foot of heavy snow. Add to that, the national average run of 100 inches of snow Tuesday -- down 20 inches from 1930's.

    BRET CADDICGIIO: Wow.

     ROBERT HORKIN: Don Dugley is the assistant meteorologist at the National Weather Service.

    BENN CADDICGIIO: Yeah. That was 1933. That was the year of storm.

    PIANY TA, HOSTATOR: This is one of the worst years of the year, just how severe the snow was.

    BRET CADDICGIIO: You know, we all predicted a golden year for Texas this year. Geez. Damn it.

    Ah, exactly what's going on.

    Hot rains in Texas, in particular, caused large numbers of homes to without electricity and stripped one-tenth-fifth of many electrical utilities in need of power. In Harris County, Texas is seeing two-and-a-half all-time-record high waters.

    Officials say they're still very early to know, but no physical evidence has been linked to high temperatures. And in the most affected state, All American County where hundreds of homes were burned due to fire, this week's freezing rain hasn't eased.

    Listen to Brian, as Nancy's mom struggles to get out of bed.

    NANDY GRATER: I'm yelling to my mom. This house is dead. This is horrible.

    NANDY GRATER: Even my mom can't get out. I don't even hear a ringing voice.

    And there haven't been five or six, though Tuesday was the largest number in Houston, per AP's count.

    PIONY GREENIN, HOSTATOR: We've got news for Houston. Texas won't be Houston, for sure, but most residents of Houston are bracing up for rising snow.

    Forecasters are searching a flood-damped area and using fire crews to clear their homes. The National Weather Service has warned residents to be prepared for this invity. Officials say their first threat is if they're homeless or can't afford shelter.

    HUGH GRANY, HIDEATOR: News from the CDC means that Texas is getting prepared for another storm. Officials at the Centers for Disease Control and the Houston Police Department are training an armored helicopter and military helicopter. The CDC is also deploying local law enforcement officers as it ramps up drug treatment and offers quick treatment to patients and families. Parts of Houston are recovering.

    GREENIN: I've been on I-10 and the middle of Houston is closed, almost closed and to eat. Temples set up to get together and hold social events, and local churches rest in tents around the streets.

    NANDY ARINVIN: I'm being prepared, especially for storms coming. Especially the plastic surgeons, my nurses, --

    AMY GRANY, HOSTATOR: Nancy Dyer has a nice job to do but she hasn't had all the help she needs. Her husband, Ayouar Talib, has trouble driving on I-35 after checking out her driver license. The road number on I-35 is --

    AMY GRANY, HOST\texttt{<|endoftext|>}
  \end{samplebox}

  \vspace{-10pt}
  \caption{Samples from \textsc{LDLM} trained on OWT (NFE: 128 and 256).}
  \label{fig:qual_sample_fix_noise_ours_owt_a}
\end{figure}

\begin{figure}[p]
  \centering
  \begin{samplebox}{\normalfont\textbf{\textsc{LDLM} (Ours), Sampling Steps: 512} \hfill \normalfont\scriptsize \textcolor{darkgray}{Gen.PPL: \textbf{28.5} \,|\, Entropy: \textbf{5.51}}}
  \scriptsize\linespread{0.85}\selectfont
      \texttt{<|endoftext|>} proud to win? We're building an even stronger defense. As long as the guys keep watching this tough game, whether it's an offensive blunder or a cold news report, it's going to be a rare and exciting day for us."

    As of Monday morning, as the Rams practiced for the first time since 2010, it hasn't turned a turn around. All season, the Rams have been 8-0 and have games against the Arizona Cowboys and Los Angeles.

    In their fourth-and-fourth seasons, they were in the NFC West. In 2009 and 2010 the Rams played games against the Green Bay Packers and Chargers. In 2011 they were at 7-4 and they were Super-10. Two years later they realized they hadn't been at 6-0 or Super-Xcellent.

    "We kicked off an really good start," said Rams coach Jim Caldwell. "I think it's going to be an unusual and exciting day for us. We had a defensive line, and we went back and did a great job getting the ball to Jared Rivers at the end of the field. We have Kenny Golladay and Sammy Jones making those great plays.

    "I think that as we enter the season, coming in back-to-back games against the Rams, our offensive line is going to be really good on the field. The line of scrimmage is going to be a lot more of an issue here in Los Angeles."

    Tracy Gordon, a first-time All-Rookie Week 1 starter, injured his ankle earlier this season. He ran for 237 yards and four touchdowns against the Rams during their 2011 Week 3 regular season.

    The Rams' offense has been dominant since notching in a 34-0 overtime loss to Philadelphia Sept. 1. They are 22th in sacks and 5.5 points behind the Washington Redskins and Atlanta Eagles in the division.

    "It goes crazy deep here, once you open the field you have the pressure," McCoy said. "You sort of lose the momentum."

    When Kaepernick asked him on the phone, he had a very small laugh.

    "Hey guys, Josh, our field-game went game-out, we couldn't beat out a bit, we couldn't run as fast as we liked," Stafford said. "And we're really looking forward to getting down a month or two. But to top things off, I'm really thankful that I'm back. They're a big, great team. I want to say thank you to the fans. They just didn't make a huge impact here in Los. I know it speaks more to me about them than I could imagine. That's the team and I want it to be."

     Contact Matt Davis at editors@washpost.com. Matt Davis is a radio reporter based in Philadelphia. He spent four years in Kansas City as quarterback. He ran 297 times and rushed for 1,000 yards and finished third in the NFL average at 39.5. Follow him on Twitter: @Matt\_davis.\texttt{<|endoftext|>} NEW DELHI: An anti-pesticine body (ARC) working committee on Monday condemned the problems involved in vaccines that caused 3,974 deaths in central India between 1995 and 2010 -- a 44 per cent drop from the previous report.

    In a statement, Suha Saeed-Chonsen, senior director of the central government, the Education and Family Health Research Board (CAV). Referring to government complaints about the lack of evidence, the working group pointed out that vaccines had caused 3,902 deaths in central India between 1995 and 2010 -- a 44 per cent decline from the previous report.

    According to CAV's annual report, 404 of the 3,651 cases of infections reported in the state were the result of vaccine management. ``We see no need to release official figures to determine when vaccines are cured,'' Saeed-Chonsen said after directing the committee's Research and Development Committee.

    On the day of the 2002 West Bengal outbreak, CAV had reported 108 people died from infections because the diagnosis process in the laboratory provided a negative link between these deaths and the polio outbreak. However, it also claimed no evidence that no record-pact-forming vaccines were administered at a laboratory that went outside prior to December 2015.

    The CAV is planning to release its review of vaccine management on Monday, but the Government Working Committee was due to release the report on Monday when it will review the evidence.

    ``The working group thus far rejected WHO's recommendation recommending that safeguards should be removed when vaccines as per the record-pague management report,'' Saeed-Chonsen said.

    The International Anti-Pesticine Organisation (IPI) is a global leader in the research and development of women's health. Saeed-Chonsen is the body dedicated to providing information on critical health.\texttt{<|endoftext|>}A 2,000-year\texttt{<|endoftext|>}
  \end{samplebox}

  \begin{samplebox}{\normalfont\textbf{\textsc{LDLM} (Ours), Sampling Steps: 1024} \hfill \normalfont\scriptsize \textcolor{darkgray}{Gen.PPL: \textbf{29.5} \,|\, Entropy: \textbf{5.50}}}
  \scriptsize\linespread{0.85}\selectfont
      \texttt{<|endoftext|>} people buying a new home or buying new homes, according to The Globe and Mail.

    ``It probably was entirely wrong to think that once the federal government had completed its key measure of labour force growth, the unemployment rate was going to drop instead,'' said Ed Bullier, a senior policy analyst at the Institute on Economic Affairs.

    ``If [abor force growth] continues so slowly over time then there may still be a sudden decline from what we've been expecting for some time.''

    [np\_storybar id=''pld1468231'' datafullwidth=''500px'']

    Just 84,000 had three years of work for a job and nearly 50,000 illegal immigrants had their wages reduced --- slightly less than half of the whole of Canada's labour force population, Bulligan said.

    The federal government approved long-term Social Security benefits to help Canada's illegal immigrants get a better way to get by in the past 10 years.

    The government also raised billions of dollars for other programs to fight immigration, such as free seaports and transit lanes.

    The government is dramatically increasing access to Social Security benefits, cutting back on promoting screening programs for illegal immigrants and working out Canadian general welfare benefits for those who illegally came to the United States with a college degree, Bullington said.

    [np\_story title=''previous'' link=''''Calgary promises to cut Canada's labour force for five years''][/relatedstrong][/np\_story]

    Social Security direct GDP grew at an 11-percent rate between 2005 and 2015, Bulligan said.

    GDP as a gross domestic product declined from 35 per cent cent between 2007-08 to 48$\tfrac{1}{2}$ per cent the previous year, and the gap continues to grow steadily among the rich and poor, Bulligan said. Many such disparities are seen as unemployment grows and opens up opportunities for people to look for a better opportunity to work.

    ``There's a great deal of variation in the economy and among people North Americans are looking for a better way to get jobs.''

    Anil Talman, an economist with the University of Toronto, said there wasn't a need to impose economic pressure on the labour force as much as a manufacturing sector downturn has led to the ``inflation crisis'' by focusing on reducing spending, Talman said.

    Garchen, who heads the Canada Economics Institute's independent think tank, said the federal government is trying to take the next steps with methodical cuts to the labour force.

    ``Who wants to punch someone out of that boat and say, `What do you do to help people buy their house or get a better job opportunity? What is the smartest way to get employed?''

    Talman said if the federal government delivered on its promises --- such as ending the Jim Crow racial profiling of several million illegal immigrants --- programs would in turn help relocate people looking for jobs.

    ``That is reality,'' he said.

    Write to Michelle.Boiz at molly.boaz@chngnews.com

    ---

    Read or Share this story: http://usat.ly/2qaQ\texttt{<|endoftext|>}Looking for news you can trust?

    Subscribe to our free newsletters.

    Two members of the Nationalist Progressive Party (CHP)'s assistant professor of orchestra at Ray University have undergone general strikes from August to September, according to the Weihi news agency (IMNA). The university faces a 30-day general strike while 200 assistant professors have been reinstated by a 30-day strike.

    Among those professors are Chen Song, deputy chair of the South Korean National Opera and Dramophone Council.

    Six of the two assistant professors have been laid off since the strike began in August, according to the CCTV news agency.

    All four of the union's other eight student members will be leaving when the strike began next week.

    Officials of the union, Kim Suhong and Dulyi Tong, said they would be appointed members of the South Korean National Opera and Gramography Council due to this August general strike.

    The chair of the union has been Dlyi Dong, an associate professor at the North Korean National Science University.

    Dlyi Ung received a degree in economics in 2010 and is a colleague of Chen Han, the former assistant professor and professor of political science at Ray University and was also an assistant professor at the education department from 2010 to 2013.

    An official for Kim Suhong and Dlyi Tong did not respond to a message seeking comment.

    Follow Ahmad Chang on Twitter @AhmadBoi\texttt{<|endoftext|>}BEIRUT, Iran -- Two foreign bank officials were caught on Thursday in a high-profile plot to topple Iran's latest regime.\texttt{<|endoftext|>}

  \end{samplebox}

  \vspace{-10pt}
  \caption{Samples from \textsc{LDLM} trained on OWT (NFE: 512 and 1024).}
  \label{fig:qual_sample_fix_noise_ours_owt_b}
\end{figure}





\end{document}